\pdfoutput=1

\documentclass[11pt]{article}

\usepackage{acl}

\usepackage{times}
\usepackage{latexsym}
\usepackage{xurl}

\usepackage{pgfplots}
\pgfplotsset{compat=newest}
\usepackage[T1]{fontenc}
\usepackage[utf8]{inputenc}
\usepackage{microtype}
\usepackage{inconsolata}
\usepackage{graphicx}
\usepackage{booktabs}
\usepackage{multirow}
\usepackage{float}
\usepackage{longtable}

\usepackage{tikz}
\usepackage{xcolor}
\usetikzlibrary{shapes.geometric, arrows.meta, positioning, fit, backgrounds, calc}
\definecolor{promptcolor}{HTML}{E8F4FD}
\definecolor{promptborder}{HTML}{2E86AB}
\definecolor{responsecolor}{HTML}{FFF3E6}
\definecolor{responseborder}{HTML}{E67E22}
\definecolor{arrowcolor}{HTML}{555555}
\definecolor{itercolor}{HTML}{F0F0F0}
\definecolor{cyanfill}{HTML}{A8E8E8}
\definecolor{cyanborder}{HTML}{40BFBF}
\definecolor{purplefill}{HTML}{C9A0DC}
\definecolor{purpleborder}{HTML}{8E44AD}
\definecolor{orangetext}{HTML}{D4880F}

\usepackage[table]{colortbl}
\usepackage{soul}
\usetikzlibrary{decorations.pathreplacing}
\usetikzlibrary{calc}
\newcommand{\tikzmark}[1]{\tikz[remember picture,overlay]\node[yshift=2pt](#1){};}

\usepackage[textwidth=0.7in]{todonotes}

\newcommand{\numDatasets}{$96$}
\newcommand{\numLanguages}{$7$}
\newcommand{\numTasks}{$5$}
\newcommand{\numModels}{$40$}

\usepackage{tcolorbox}
\usepackage{amsmath,amssymb}
\usepackage{makecell}
\usepackage{authblk}

\title{STEB: Style Text Embedding Benchmark}

\author[1]{\bf Rafael Rivera Soto}
\author[2]{\bf Anna Wegmann}
\author[1]{\bf Cristina Aggazzotti}
\affil[1]{Johns Hopkins University}
\affil[2]{Utrecht University}
\affil[]{\texttt{rafaelriverasoto@jhu.edu, a.m.wegmann@uu.nl, caggazz1@jhu.edu}}

\begin{document}
\maketitle
\begin{abstract}
While semantic embeddings are rigorously evaluated on the 
Massive Text Embedding Benchmark, the evaluation of style embeddings remains fragmented, with each work relying on their own set of tasks and datasets.
To bridge this gap, we introduce the Style Text Embedding Benchmark, a comprehensive open-source benchmark intended to standardize the evaluation of style embeddings.
\texttt{STEB} encompasses \numDatasets{} datasets across \numLanguages{} languages, spanning applications such as authorship verification, authorship retrieval, AI-text detection, probing of linguistic features, and others.
We find that semantic embeddings consistently fail in stylistic tasks, and that there is no style embedding that is universally superior across all tasks evaluated.
We open-source the \texttt{STEB} code base at: \url{https://github.com/rrivera1849/STEB}.
\end{abstract}

\section{Introduction} \label{introduction}

\begin{figure}[!t]
    \centering
    \includegraphics[width=\columnwidth, scale=0.8]{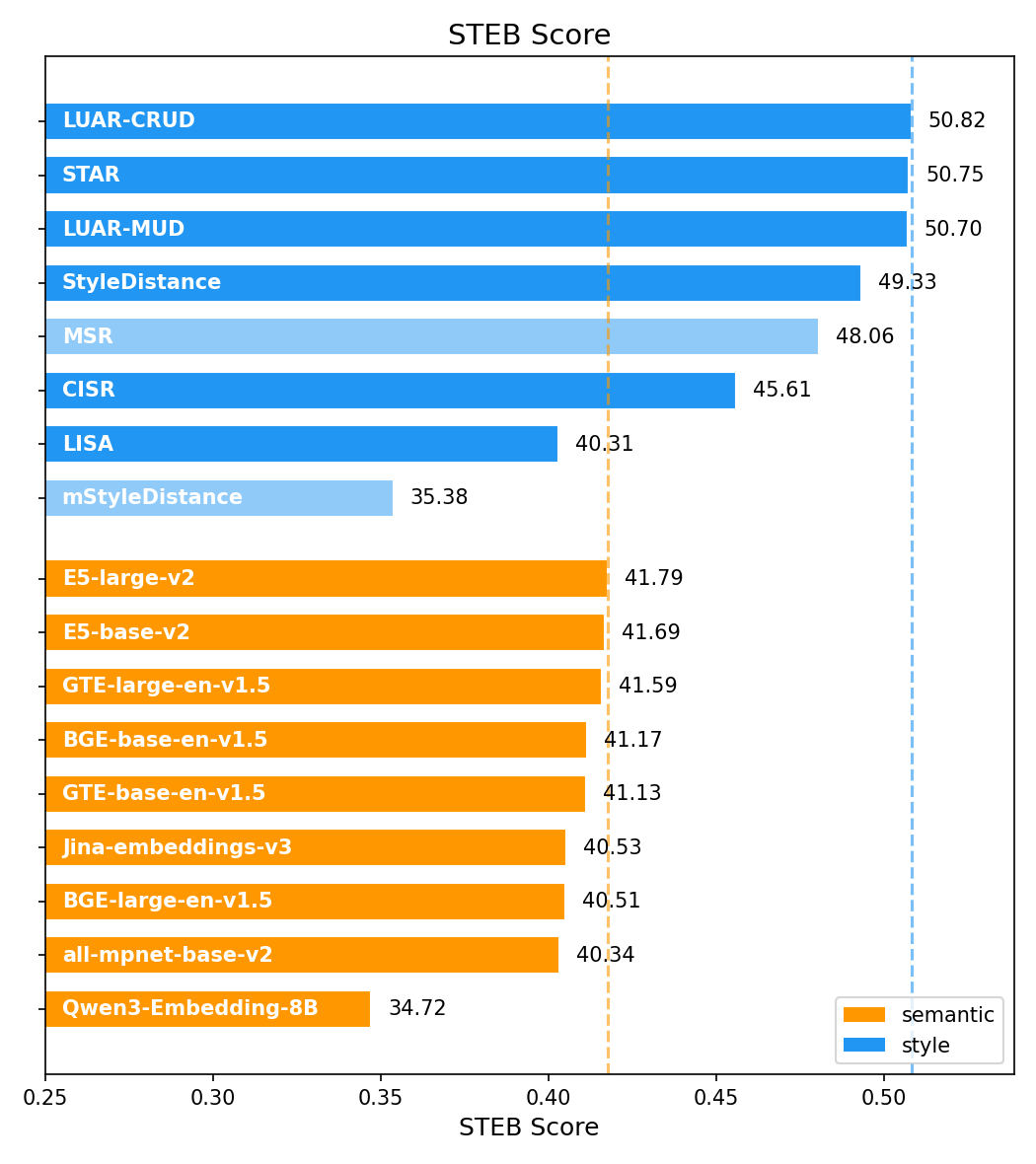}
    \caption{\textbf{\texttt{STEB} Score by model category}
    Style embeddings (blue) score above general-purpose semantic models (orange) on stylistic tasks; dashed lines mark the best model in each category.
    \emph{Qwen3-Embedding-8B, a top-5 MTEB model, ranks poorly on \texttt{STEB}.}}
    \label{fig:model_ranking}\vspace{-.4cm}
\end{figure}

Representation learning has driven progress in NLP in the last decade, including on information retrieval, clustering, classification, 
and semantic search.
While such models like SBERT-models~\citep{reimers_sentence-bert_2019}, E5~\citep{wang2024textembeddingsweaklysupervisedcontrastive}, LLM2Vec~\citep{behnamghader2024llmvec}, and EmbeddingGemma~\citep{vera2025embeddinggemmapowerfullightweighttext} focus on capturing semantic meaning, a more disregarded parallel field focuses on the \emph{style} of text.\footnote{Semantic meaning (or content) and style are hard to disentangle and might not be fully disjoint \cite{
wegmann_survey_2026}.
}
Style representations have proven valuable in applications such as authorship attribution~(AA), style transfer, and AI-generated text detection~\citep{horvitz_paraguide_2024, horvitz_tinystyler_2024, khan2024learninggeneratetextarbitrary, kim_leveraging_2025, rivera_soto_learning_2021, rivera_soto_few-shot_2023, wegmann-etal-2022-author}, and might support the development of more style-aware LLMs \cite{wegmann_survey_2026}. 

Currently, evaluation approaches for style embeddings are inconsistent.
Works vary considerably in the tasks and datasets they evaluate on and in evaluation protocol decisions like preprocessing, encoded text length, and documents per embedding.
For example, %
LUAR~\citep{rivera_soto_learning_2021} evaluates only on authorship retrieval, STAR adds author clustering~\citep{10.1016/j.knosys.2024.111867}, and LISA~\citep{patel_learning_2023} uses the STEL evaluation framework~\citep{wegmann-nguyen-2021-capture,wegmann-etal-2022-author}.
The Massive Text Embedding Benchmark (MTEB)~\citep{muennighoff_mteb_2023}, positioned as a general-purpose text embedding benchmark, offers no alternative as it  includes no style-specific tasks, and many of its leading models~\citep{bai2023qwentechnicalreport,lee2025nvembed,li2023generaltextembeddingsmultistage,wang2024textembeddingsweaklysupervisedcontrastive,bge_embedding} are trained on objectives such as semantic similarity, information retrieval, and natural language inference (NLI) that %
benefit less from %
stylistic features (see~\autoref{fig:model_ranking}).
As a result, cross-work comparisons are unreliable, and progress on style embeddings is hard to quantify.

We introduce the open-source Style Text Embedding Benchmark (\texttt{STEB}).
\texttt{STEB} consists of \numDatasets{} datasets across \numLanguages{} languages, organized into \numTasks{} evaluation tasks (clustering, pair classification, order alignment, retrieval, and probing) under fixed evaluation protocols (e.g., metrics, long document handling) covering applications like AI-text detection (ATD), authorship verification (AV) and authorship retrieval (AR).
\texttt{STEB} reports two complementary scores. 
The \textbf{operational} score reflects the datasets and tasks that the field has built to study style. It inherits the emphasis of the field---most prominently AA work---without committing to an a priori definition. 
The \textbf{definitional} score instead reweighs results according to the style definition proposed in \citet{wegmann_survey_2026}.
We evaluate \numModels{} models spanning style embeddings, general-purpose semantic embeddings, masked language models (MLMs), causal language models (CLMs) and non-neural baselines.
We find that no single model dominates across \texttt{STEB}. Instead, the best representation depends on whether the goal is strong performance on a specific downstream application or broad coverage of stylistic attributes, as well as on the degree to which the task is entangled with semantics. 
Notably, recent semantic embeddings that lead MTEB (e.g., Qwen-Embedding-8B) perform substantially worse than specialized style representations, highlighting a gap in MTEB’s evaluation of stylistic tasks. 
Furthermore, re-evaluating a prior multilingual setup under a different protocol for fairly handling long texts produces substantially different rankings, reinforcing the importance of standardized evaluation procedures. 
We also find that off-the-shelf MLMs perform surprisingly well at capturing linguistic features and remain competitive on AV and AR, suggesting promising directions for future research. 
Overall, we hope that \texttt{STEB} serves as a ``yardstick'' for measuring progress toward better style representations.

\paragraph{Why not just use LLM prompting?}
While generative models can solve an increasing number of problems, %
text representations remain highly relevant and widely used in practice \citep{enevoldsen_mmteb_2025, warner_smarter_2025} for several reasons: 
(i) Text representations are more efficient: state-of-the-art encoder models are smaller (millions vs. billions of parameters) and only use a single forward pass, whereas generative models incur cost proportional to the output length; (ii) they are competitive on discriminative tasks, matching or outperforming much larger generative models \citep{warner_smarter_2025}; and (iii) they scale much better to larger document pools for applications like RAG \citep{ram_-context_2023} or clustering, and might reduce hallucinations \citep{gao_retrieval-augmented_2024}.
We demonstrate that LUAR-CRUD~\citep{rivera_soto_learning_2021} clearly beats GPT-5.2 on a small AR task,  %
while using $\gg$750$\times$ fewer FLOPs (\autoref{tab:retrieval-llm}; full setup in App.~\ref{app:small-retrieval}).

\begin{table}[h!]
\centering
\small
\setlength{\tabcolsep}{4pt}
\begin{tabular}{lccccc}
\toprule
\textbf{Model} & \textbf{R@1} & \textbf{R@8} 
& \textbf{MRR} & \textbf{TFLOPs} & \textbf{USD} \\  
\midrule
{LUAR-CRUD} & \textbf{83.0} & \textbf{95.0} 
    & \textbf{87.8} & $\approx32$ & 0 \\
GPT-5.2 & 59.0 & 69.0 %
    & 63.4 & $\gg24$k & 22 \\
\bottomrule
\end{tabular} 
\caption{{Embeddings outperform LLMs in AR at a fraction of the compute.}
The GPT-5.2 FLOP estimate uses 1 billion active parameters as a conservative lower bound.}
\vspace{-.4cm}
\end{table}\label{tab:retrieval-llm}

\section{Related works}

\paragraph{Benchmarks for style representations}
Evaluation of style representations has mostly focused on authorship tasks \cite{wegmann_survey_2026}.
The longest-running effort in this space is the PAN shared task series at CLEF, which has run continuously since the late 2000s and describes itself as covering ``digital text forensics and stylometry''.\footnote{\url{https://pan.webis.de/}}
Early editions established cross-domain AA and AV as standard tasks~\citep{Juola2013OverviewOT,kestemont_overview_nodate-2,pan2014,stamatatos_overview_2015,stamatatos_overview_2018}, while later editions broadened to author profiling, style change detection, multi-author writing style analysis, hate-speech spreader profiling, and ATD~\citep{pan2024,pan2020,pan2021,pan2025,pan2026_preprint,pan2023}.
Most of the style embeddings we evaluate, evaluate on at least one PAN split. 
For AR, LUAR~\citep{rivera_soto_learning_2021} introduces its own evaluation splits, which have since continued to be used (e.g.,~\citet{man2026explainabledisentangledrepresentationlearning}). 
Few evaluation approaches have used more theoretical definition-based understandings. %
For example, STEL~\citep{wegmann-nguyen-2021-capture} introduces a parallel-text ranking evaluation over a small set of style dimensions, and STEL-or-Content~\citep{wegmann-etal-2022-author} extends it with a content-controlled variant that penalizes models reliant upon semantics.
\citet{patel_styledistance_2025} further extends STEL by prompting LLMs to generate synthetic instances across $40$ stylistic features.

\section{The Style Text Embedding Benchmark} \label{sec:steb}

\texttt{STEB} consists of \numDatasets{} datasets across \numLanguages{} languages, organized into \numTasks{} evaluation tasks.
In Section \ref{sec:tasks}, we outline these tasks, define the metrics used for scoring each, and discuss how each serves to evaluate the quality of style embeddings. In \autoref{sec:STEB-stats}, we provide information about the included datasets and their properties. In \autoref{sec:STEB-score}, we explain the operational and the definitional STEB score. In \autoref{sec:extending-steb}, we mention principles that we considered, and should generally be considered, when extending \texttt{STEB}.

\subsection{Tasks and evaluation}\label{sec:tasks}
See \autoref{tab:STEB-examples} for examples of all task types in \texttt{STEB}.

\paragraph{Pair classification}
The goal of pair classification is to determine whether two input texts share the same label (e.g., same author in AV).
To perform this, we embed both inputs and compute their cosine similarity, expecting high similarity between same-label pairs and low similarity otherwise. 
We implement two variants: (1) \textbf{All-to-All}, where every sample in the set is compared against every other sample providing an exhaustive evaluation, and (2) \textbf{Predefined}, where the pairs are restricted to a predefined list of pairs.
This latter setting is critical for controlling for confounding variables (e.g., enforcing cross-topic pairs).
We report the area under the receiver operating curve (AUROC). %

\paragraph{Clustering}
While semantic embeddings are optimized to cluster documents by topic, style embeddings should induce clusters corresponding to authorship~\citep{andrews-bishop-2019-learning}, register (e.g.,\ formality vs.\ informality)~\citep{patel_styledistance_2025}, and LM provenance~\citep{rivera_soto_few-shot_2023}.
We evaluate whether style embeddings are capable of clustering text samples across style-relevant labels.
Following the MTEB~\citep{muennighoff_mteb_2023} protocol, we use a mini-batch $k$-means algorithm with a batch size of $32$ and $k$ equal to the number of ground-truth labels.
We report the V-measure~\citep{rosenberg-hirschberg-2007-v}. %

\paragraph{Authorship retrieval}
AR has emerged as a primary application of style embeddings~\citep{agarwal_cross-genre_2025,fincke_separating_2024,kim_leveraging_2025, man_explainable_2026,rivera_soto_learning_2021}.
Unlike traditional information retrieval, which optimizes for semantic relevance between a query and a document, AR aims to identify other texts written by a specific author from within a large candidate pool.
We use standard retrieval benchmarks where topic and authorship may correlate but are not the sole signal.
We report the mean reciprocal rank (MRR). %

\paragraph{Order alignment}
Order alignment evaluates whether a style embedding ranks a set of texts in the same stylistic order as a reference set.
In its simplest form, this corresponds to the STEL task introduced by \citet{wegmann-nguyen-2021-capture}, which uses parallel texts.
A pair of texts has to be aligned to the order of another base pair of texts using stylistic information (see App.~\autoref{fig:order-alignment-example}).
Every order alignment task also has a \textbf{distractor} variant.
Here, the unordered set includes one additional text that shares the same topic as the ordered texts but stylistically unrelated to any of them, a model that relies on semantic features more than style will mis-align it.
In a setup with a one element ordered set, and with a two element unordered set including one distractor, this is equivalent to the STEL-or-Content task \cite{wegmann-etal-2022-author}.
We report two accuracies, one for the original and one for the distractor variant.
To achieve a high score in this task, an embedding \emph{must} capture features other than semantics, and in the distractor variant it must ignore semantic similarity altogether.

\paragraph{Probing}
The probing task assesses which linguistic features are encoded in an embedding by training a linear classifier on top of frozen representations~\citep{conneau_what_2018}.
We extract foundation-level linguistic features using the LFTK toolkit~\citep{lee_lftk_2023}, excluding features that may correlate with semantics (e.g., entity counts).
Each continuous feature is discretized into five quantile bins, and the train/validation/test splits are balanced across the discretized labels.
We then train a logistic regression probe on the frozen embeddings, selecting the L2 regularization strength from $\{10^{-5}, 10^{-4}, 10^{-3}, 10^{-2}\}$ via a validation set, and report test accuracy averaged across all features.

\subsection{\texttt{STEB} datasets} \label{sec:STEB-stats}

An overview of all datasets can be found in \autoref{sec:dataset_details}.
The added datasets can generally be grouped into 6 categories: those that target capabilities with respect to single linguistic features (7 datasets), authorial styles (58), dialectal styles (3), registers (12), genres (4), historical styles (3) and demographics (9).

\subsection{\texttt{STEB} Scores} \label{sec:STEB-score}

With \texttt{STEB}, we evaluate models on the most common datasets and tasks that the field has built to study style. We aggregate the results differently for the operational score and the definitional score.

\paragraph{Operational \texttt{STEB} Score} %
While we do not align the operational score with a definition, we take care to not overweigh groups of correlated datasets. %
We automatically discover redundancies following \citet{olmo2026olmo3} using the Spearman correlation between the model rankings on each dataset and apply hierarchical clustering (Ward linkage with a $0.5$ distance threshold, cf.~\citealp{Ward01031963}). 
Scores are macro-averaged within each %
and then across clusters.

\paragraph{Definitional \texttt{STEB} Score} To contrast to the operational view of style (cf.~\autoref{introduction}), we also evaluate embeddings based on an a priori definition of style, namely the  definition introduced in \citet{wegmann_survey_2026}. We evaluate whether embeddings (i) encode linguistic features, (ii) capture the stylistic patterns of different objects of study (e.g., dialect, genre, idiolect), and (iii) are more sensitive to stylistic than content information. %
See App.~\autoref{tab:dataset-style-score} for an overview of the dataset-to-cluster assignment.

\subsection{Design principles} \label{sec:extending-steb}

We provide code and datasets with \texttt{STEB} that are meant to be extended by the community. We emphasize the principles on which we designed \texttt{STEB} and that we recommend the community to follow: 

\paragraph{Tasks where style matters}
Fundamentally, \texttt{STEB} targets scenarios where the semantics of a text is neither the sole nor strongest signal.
Where possible, we enforce topic-controlled settings that penalize models that rely on any semantic shortcuts (e.g.,~\autoref{tab:per-application-clusters}). 
As such, each \texttt{STEB} dataset should benefit from representations that capture stylistic information and should be more difficult for models that heavily rely on semantic information.

\paragraph{Length control}
To preserve ecological validity~\citep{Vries2020TowardsEV}, we evaluate models on the full text length inherent to each dataset.
This approach preserves the natural length distributions between domains, for example, retaining the verbosity of blog posts versus the brevity of Reddit comments. 
For models with fixed context windows (e.g., $512$ tokens), we employ a chunk-and-pool strategy where inputs exceeding the limit are segmented in sentence-boundary-aware chunks.
We embed each chunk independently and apply mean pooling to derive a single document representation (see~\autoref{fig:chunking} for a visual of the approach).

\paragraph{Reproducibility and extensibility} 
For reproducibility, all stochastic steps in evaluation (e.g., $k$-means initialization, probe training) use fixed random seeds.
To facilitate community adoption, \texttt{STEB} is open-source and easy to extend.
We welcome contributions from the community in adding new tasks and datasets.

\section{Models}\label{sec:models}

We test non-neural (\autoref{sec:non-neural}) as well as automatically-learned representations (\autoref{sec:embeddings}).

\subsection{Non-neural representations} \label{sec:non-neural}
We test various representations using predefined, linguistically motivated features. %
Note that the strongest non-neural representation generally combines several of these features and tailors the set to the specific dataset or task \cite{wegmann_survey_2026}.

\paragraph{Function word frequencies}
Function words (i.e., stop words) like determiners (\emph{the}) %
and pronouns (\emph{you}) %
are useful across text analyses, such as AA \cite{kestemont2014,mostellerwallace1963}, genre/text type classification \cite{venglarova2024}, and deception detection \cite{liu2012}. We count the frequencies of 390 function words and 69 function phrases (\emph{instead of}), %
saving these counts as a vector per text \cite{aggazzotti2025-stylospeaker}. %

\paragraph{TF-IDF-weighted n-grams} N-grams---se\-quen\-ces of $n$ consecutive units (i.e., characters, tokens, part-of-speech (POS) tags)---have long been used for text analysis, especially for AA \cite{houvardasstamatatos2006, peng2003}. Following others (e.g., \citealp{weerasinghe2020}), we weight the n-grams by their Term Frequency-Inverse Document Frequency (TF-IDF) and fit the vectorizer to a 10 billion token sample of FineWeb \cite{penedo2024}; see \autoref{app:tfidf} for more details.\footnote{Note that n-grams can capture topic as well as style signals, depending on the n-gram unit and their frequency.
}

\paragraph{Stylometric features}
There is no agreed upon fixed set of best stylometric features \cite{juola2006,nini2023}. There are perhaps equally as many feature extraction tools for various use cases. Due to its popularity in the NLP community, we use LFTK \cite{leelee2023-lftk} and combine its surface and POS features, which we found to perform best across datasets while remaining fast. %

\paragraph{NeuroBiber}
NeuroBiber \cite{alkiek_neurobiber_2025} is a transformer-based system that %
extracts 96 syntactic and lexical ``Biber features'', such as function words, subordination types, and tenses \cite{biber1988,biberconrad2009}. 

\begin{table*}[!t]
\centering
\small
\setlength{\tabcolsep}{4pt}
\begin{tabular}{cl cccccc c}
\toprule
& \textbf{Model} & \textbf{Clustering} & \textbf{All-to-All} & \textbf{Predefined} & \textbf{Order Align.} & \textbf{Retrieval} & \textbf{Probing} & \textbf{\texttt{STEB} score} \\
& Num. Datasets & 53 & 24 & 26 & 11 & 4 & 4 & 122 \\ 
\midrule
\multirow{8}{*}{\rotatebox[origin=c]{90}{\shortstack{\emph{Style-}\\\emph{specific}}}}
& LUAR-MUD & 20.03 & 62.33 & 72.33 & 20.81 & \underline{75.67} & 53.04 & 50.70 \\
& LUAR-CRUD & 19.65 & 61.00 & \underline{73.12} & 19.80 & \textbf{77.64} & 53.75 & \textbf{50.82} \\
& StyleDistance & 18.46 & 59.82 & 66.21 & \textbf{44.02} & 49.61 & \underline{57.87} & 49.33 \\
& mStyleDistance & 7.35 & 53.83 & 53.71 & 31.38 & 20.04 & 45.96 & 35.38 \\
& MSR & 16.67 & \underline{64.34} & 71.01 & 21.85 & 63.49 & 51.01 & 48.06 \\
& CISR & 15.28 & 59.71 & 66.43 & \underline{33.81} & 45.87 & 52.54 & 45.61 \\
& STAR & \textbf{25.92} & \textbf{64.56} & \textbf{73.28} & 23.93 & 66.96 & 49.86 & \underline{50.75} \\
& LISA & 12.74 & 60.91 & 62.40 & 16.48 & 42.96 & 46.38 & 40.31 \\
\midrule
\multirow{3}{*}{\rotatebox[origin=c]{90}{\shortstack{\emph{Sem-}\\\emph{antic}}}}
& all-mpnet-base-v2 & 8.48 & 60.41 & 63.87 & 14.62 & 48.99 & 45.64 & 40.34 \\
& E5-large-v2 & 11.06 & 60.20 & 64.70 & 15.82 & 49.91 & 49.03 & 41.79 \\
& Qwen3-Embedding-8B & 2.67 & 51.61 & 52.79 & 23.11 & 35.17 & 42.96 & 34.72 \\
\midrule
\multirow{4}{*}{\rotatebox[origin=c]{90}{\emph{MLM}}}
& BERT-large-cased & 12.02 & 60.99 & 67.11 & 21.57 & 58.88 & 55.16 & 45.96 \\
& RoBERTa-large & 13.21 & 60.97 & 68.79 & 20.71 & 61.07 & \textbf{57.92} & 47.11 \\
& DeBERTa-v3-large & \underline{21.44} & 59.37 & 70.45 & 24.23 & 63.24 & 53.41 & 48.69 \\
& ModernBERT-large & 16.58 & 60.19 & 65.38 & 20.56 & 52.17 & 54.06 & 44.82 \\
\midrule
\multirow{2}{*}{\rotatebox[origin=c]{90}{\shortstack{\emph{CLM}}}}
& OPT-1.3B & 3.83 & 52.40 & 55.59 & 24.08 & 39.07 & 45.82 & 36.80 \\
& Qwen3.5-4B-Base & 4.23 & 52.80 & 57.41 & 24.88 & 37.15 & 43.02 & 36.58 \\
\midrule
\multirow{4}{*}{\rotatebox[origin=c]{90}{\shortstack{\emph{Non-}\\\emph{neural}}}}
& Function words & 7.17 & 55.17 & 60.88 & 14.51 & 43.14 & 37.98 & 36.47 \\
& TFIDF n-grams & 6.78 & 57.18 & 64.45 & 13.72 & 50.95 & 55.74 & 41.47 \\
& Stylometric & 9.67 & 56.88 & 60.48 & 25.87 & 39.52 & 53.30 & 40.95 \\
& NeuroBiber & 8.24 & 55.10 & 57.27 & 14.13 & 9.06 & 51.64 & 32.57 \\
\bottomrule
\end{tabular}
\caption{Overall \texttt{STEB} (operational) results ($\times 100$). \textbf{Bold} = best, \underline{underline} = 2nd best per column. Metrics: V-measure (Clustering), AUC (All-to-All and Predefined Pair Classification), distractor accuracy (Order Alignment), MRR (Retrieval), and average accuracy (Probing). Scores are macro-averaged across automatically discovered dataset clusters within each task, then averaged across tasks to produce the overall \texttt{STEB} score.
Num. Datasets = datasets contributing to each task; a dataset may contribute to more than one task. %
The full table is in App.~\autoref{tab:overall-results-full}.
}
\label{tab:overall-results}%
\end{table*}

\begin{table*}[!t]
\centering
\small
\setlength{\tabcolsep}{4pt}
\begin{tabular}{cl cccccc c c c c}
\toprule
& & \multicolumn{7}{c}{\textbf{Object of Study}} & \textbf{Ling. Feat.} & \textbf{Content-Ind.} & \textbf{Avg.} \\
\cmidrule(lr){3-9} \cmidrule(lr){10-10} \cmidrule(lr){11-11}
& \textbf{Model} & \textbf{Genre} & \textbf{Register} & \textbf{Time} & \textbf{Demo.} & \textbf{Dialect} & \textbf{Idiolect} & \textbf{Avg.} &  & \textbf{Order Al.} & \\
& Num. Datasets & 4 & 7 & 3 & 3 & 3 & 29 & 49 & 7 & 11 & 67 \\
\midrule
\multirow{5}{*}{\rotatebox[origin=c]{90}{\shortstack{\emph{Top 5}}}} 
& StyleDistance & 62.84 & 40.01 & 46.21 & 33.67 & \textbf{56.91} & 60.03 & 49.94 & 74.02 & \underline{44.23} & \textbf{56.06} \\
& StyleDist. Synth. & 55.45 & 38.81 & 43.27 & 33.13 & \underline{55.64} & 45.26 & 45.26 & 70.87 & \textbf{46.10} & \underline{54.08} \\
& CISR & 62.43 & 36.09 & 41.52 & 33.03 & 54.76 & 58.72 & 47.76 & 65.85 & 32.31 & 48.64 \\
& DeBERTa large %
    & 63.26 & 38.25 & \underline{46.94} & 35.37 & 43.37 & 68.26 & 49.24 & 69.56 & 18.72 & 45.84 \\
& RoBERTa-base & 63.79 & \textbf{42.52} & 45.94 & 32.84 & 40.16 & 59.83 & 47.51 & \textbf{77.95} & 10.27 & 45.24 \\
\midrule
\multirow{5}{*}{\rotatebox[origin=c]{90}{\shortstack{\emph{Style emb.}}}} 
& STAR & \textbf{68.68} & 39.79 & \textbf{47.25} & \textbf{40.99} & 49.39 & 72.24 & \textbf{53.06} & 67.66 & 14.08 & 44.93 \\
& LUAR-MUD & \underline{66.67} & 38.72 & 42.14 & 35.53 & 44.30 & \underline{76.11} & \underline{50.58} & 70.50 & 11.29 & 44.12 \\
& mStyleDistance & 49.41 & 32.83 & 35.04 & 32.25 & 42.83 & 37.23 & 38.27 & 56.17 & 38.32 & 44.25 \\
& LUAR-CRUD & 65.80 & 38.14 & 41.30 & 35.32 & 41.52 & \textbf{76.99} & 49.85 & 70.37 & 10.57 & 43.59 \\
& MSR & 66.39 & 39.08 & 41.98 & \underline{37.05} & 41.52 & 68.69 & 49.12 & 66.52 & 12.35 & 42.66 \\
& LISA & 56.89 & 37.95 & 37.38 & 32.95 & 27.19 & 54.00 & 41.06 & 64.02 & 7.10 & 37.39 \\
\bottomrule
\end{tabular}
\caption{Top 5 models for \texttt{STEB} Score (definitional). %
Attributes are grouped into three  clusters after \citet{wegmann_survey_2026} and averaged: Object of Study (Genre, Style, Time, Demographics, Dialect, Idiolect), Linguistic Features, and Content-Independence. See App.~\autoref{tab:dataset-style-score} for selected datasets and App.~\autoref{tab:attribute-clusters-all} for full results.
\label{tab:attribute-clusters}}
\end{table*}

\subsection{Neural models} \label{sec:embeddings}
We benchmark models with state-of-the-art results on various style and semantic embedding tasks, grouped into four families: style embeddings, semantic embeddings, MLMs, and CLMs.

\paragraph{Style embeddings}
Most style embedding models we evaluate are trained contrastively, differing primarily in how they construct positive and negative pairs.
LUAR~\citep{rivera_soto_learning_2021} and STAR~\citep{10.1016/j.knosys.2024.111867} treat texts by the same author as positives and different authors as negatives, learning authorship representations from Reddit and various social media sources, respectively.
CISR~\citep{wegmann-etal-2022-author} further constrains positives to be same-author but \emph{different-topic} pairs, encouraging content-independent style representations.
StyleDistance~\citep{patel_styledistance_2025} constructs synthetic hard-positives and hard-negatives designed to isolate stylistic correlations from topical ones.
mStyleDistance~\citep{qiu_mstyledistance_2025} extends this approach to the multilingual setting.
MSR~\citep{kim-etal-2025-leveraging} learns multilingual authorship embeddings using probabilistic content masking to suppress content features and language-aware batching to reduce cross-lingual easy-negatives.
LISA~\citep{patel_learning_2023} departs from purely contrastive training, learning interpretable style embeddings via a linear projection over an EncT5 encoder.

\paragraph{Semantic embeddings}
These models are trained for general-purpose text similarity, primarily on semantic retrieval and NLI data.
We include models like 
all-mpnet-base-v2~\citep{reimers_sentence-bert_2019} %
that perform well according to the semantic textual similarity evaluations and various models that perform(ed) well according to MTEB \cite{muennighoff_mteb_2023}: 
Qwen3-Embedding-8B~\citep{qwen3embedding}, %
GTE~\citep{li2023generaltextembeddingsmultistage}, %
E5~\citep{wang2024textembeddingsweaklysupervisedcontrastive}, 
BGE~\citep{bge_embedding}, %
and Jina Embeddings v3~\citep{10.1007/978-3-031-88720-8_21}. %

\paragraph{Masked language models (MLMs)}
We include encoder-only transformers pre-trained with masked language modeling and without embedding-specific fine-tuning.
Specifically, we include RoBERTa~\citep{liu2019robertarobustlyoptimizedbert}, %
DeBERTa-v3~\citep{he2023debertav}, %
and ModernBERT~\citep{warner_smarter_2025}. %
Since these models do not produce a single vector, we obtain embeddings by mean-pooling the token-level hidden states from the last layer, weighted by the attention mask.

\paragraph{Causal language models (CLMs)}
Auto-regressive models are pretrained with next-token prediction and have no embedding-specific training. We include them to assess whether large-scale causal language modeling implicitly captures stylistic features.
We include GPT-2 XL~\citep{radford_language_2019}, %
OPT-1.3B~\citep{zhang2022optopenpretrainedtransformer} %
and the Qwen family at multiple scales. %
We use the hidden state of the last non-padding token as the embedding, as it is the only position with full sequence context.

\section{Results}

We discuss operational and definitional \texttt{STEB} results (\autoref{sec:overall-results}) and provide a per-application (\autoref{sec:per-application-analysis}) and a multilingual (\autoref{sec:multilingual-analysis}) analysis; all but the last are restricted to English.

\subsection{Overall Results}
\label{sec:overall-results}

\texttt{STEB} (operational) results are in~\autoref{tab:overall-results} and
selected \texttt{STEB} (definitional) results in \autoref{tab:attribute-clusters}; full results for all \numModels{} models are in App.~\autoref{tab:overall-results-full} and \autoref{tab:attribute-clusters-all}. %

\begin{table*}[!t]
\centering
\footnotesize
\setlength{\tabcolsep}{4pt}
\begin{tabular}{cl cc cccc cc}
\toprule
& \textbf{Model} & \textbf{ATD} & \textbf{ATD (Adv.)} & \multicolumn{4}{c}{\textbf{Authorship Verification}} & \textbf{AR} & \textbf{Avg.} \\
\cmidrule(lr){4-7}
& &  &  & \textbf{Overall} & \textbf{Easy} & \textbf{Medium} & \textbf{Hard} &  &  \\
\midrule
\multirow{5}{*}{\rotatebox[origin=c]{90}{\shortstack{\emph{Top 5}}}} 
& deberta-v3-large & \textbf{36.03} & \textbf{100.00} & 73.29 & 77.68 & 62.92 & 55.87 & 63.24 & \textbf{68.14} \\
& STAR & 22.79 & \textbf{100.00} & \textbf{77.52} & 88.37 & \textbf{76.02} & 58.26 & 66.96 & \underline{66.82} \\
& LUAR-MUD & 28.20 & 75.01 & \underline{76.54} & 86.84 & 73.62 & 61.10 & \underline{75.67} & 63.85 \\
& deberta-v3-base & 29.55 & \underline{99.24} & 69.40 & 73.46 & 59.80 & 54.75 & 55.17 & 63.34 \\
& LUAR-CRUD & 25.20 & 72.61 & 76.35 & 86.60 & \underline{75.92} & 61.23 & \textbf{77.64} & 62.95 \\
\midrule
\multirow{6}{*}{\rotatebox[origin=c]{90}{\shortstack{\emph{Style}\\\emph{Embeddings}}}}
& StyleDistance & \underline{30.10} & 70.08 & 70.44 & 75.52 & 69.17 & \underline{62.18} & 49.61 & 55.06 \\
& mStyleDistance & 15.85 & 5.48 & 54.41 & 60.49 & 49.63 & 50.53 & 20.04 & 23.95 \\
& StyleDistance (Synthetic) & 22.76 & 44.03 & 59.00 & 64.82 & 55.28 & 52.62 & 31.53 & 39.33 \\
& MSR & 17.38 & 0.12 & 73.88 & \textbf{88.92} & 73.43 & 55.99 & 63.49 & 38.72 \\
& CISR & 22.78 & 64.33 & 71.57 & 79.39 & 70.74 & \textbf{65.56} & 45.87 & 51.14 \\
& LISA & 11.61 & 6.06 & 65.03 & 84.41 & 66.71 & 47.86 & 42.96 & 31.41 \\
\bottomrule
\end{tabular}
\caption{Top 5 models across application clusters, plus the remaining style embeddings. 
ATD = AI-Text Detection, AR = Authorship Retrieval. 
\textbf{Authorship Verification}: \emph{Overall} aggregates a superset of datasets (PAN13--15, PAN20--21, Enron, and the PAN22--26 style-change benchmarks), while \emph{Easy}/\emph{Medium}/\emph{Hard} are computed over the PAN22--26 style-change subset only---they do not average to the \emph{Overall} column. 
To avoid double-counting, only \emph{Overall} contributes to \emph{Avg.} 
\textbf{Bold} = best, \underline{underline} = 2nd best per column (across all models). Each column reports the macro-averaged score ($\times 100$) within the corresponding manual cluster. \label{tab:per-application-clusters}}
\end{table*}

\paragraph{No single winner across task categories and style definitions}
There is no consistent winner across categories, for neither \texttt{STEB} (definitional) nor \texttt{STEB} (operational). 
While StyleDistance wins for the definitional, STAR, LUAR-CRUD, and LUAR-MUD score similarly for the operational score.
STAR's advantage is in clustering and predefined pair classification, while it falls behind on retrieval, where LUAR-CRUD leads by over $10$ points, and order alignment, where StyleDistance leads by nearly $19$ points. 
Models like STAR and LUAR might profit from content-entangled features for different objects of study that are punished more heavily for the definitional STEB score.
These patterns confirm previous observations that style is conceptualized differently across the literature, and that different training and evaluation objectives---motivated by different style definitions---result in different strengths and weaknesses across downstream tasks~\citep{wegmann_survey_2026}.

\paragraph{Style embeddings consistently outperform semantic embeddings.}
Every style-specific model except LISA and mStyleDistance outperform the best general-purpose semantic embedding on both \texttt{STEB} scores.
This disparity is especially pronounced in order alignment, AR, and predefined pair classification.
Semantic embeddings are optimized on objectives that favor matching texts primarily by semantic content, and the gap on \texttt{STEB} suggests style features are not well captured by semantic objectives alone.
These findings underscore the need for a dedicated style benchmark, as semantic-oriented evaluation fails to capture the capabilities of style-specific models.

\paragraph{MLM pre-training captures stylistic information.}
All MLMs we evaluate outperform every semantic embedding and every causal LM on both \texttt{STEB} scores, despite having no embedding-specific fine-tuning.
DeBERTa-v3-large outperforms some style embeddings on retrieval and matches style-specific models on authorship verification, \emph{without} any author-specific training (\autoref{tab:overall-results}). RoBERTa leads over style embeddings in sensitivity to linguistic features (probing column of~\autoref{tab:overall-results}; linguistic-feature column of~\autoref{tab:attribute-clusters}), suggesting that style-specific fine-tuning can come at the cost of linguistic feature sensitivity.
Joint training on MLM and style objectives may help recover this sensitivity in style-specific models.
The newer ModernBERT variants (only shown in App.~\autoref{tab:overall-results-full} \&~\autoref{tab:attribute-clusters-all}) underperform, %
suggesting that more recent MLM recipes have shifted away from preserving linguistic feature information.
Together, these results suggest that new MLM objectives are a promising direction for future research.

\paragraph{Non-neural models are competitive on certain tasks.} 
TFIDF n-grams is the strongest non-neural model, followed closely by Stylometric. Both generally outperform causal LMs and perform similarly to general-purpose sentence embedders, but lag behind style-specific and masked models. They perform best on tasks like probing and predefined classification, but degrade on tasks like clustering and retrieval that often require richer representations. Despite TFIDF n-grams being effectively a bag-of-words model, it nearly matches sophisticated sentence embedders. Also, although the literature has shown no general-purpose best set of stylometric features, using surface and POS tag features does fairly well across datasets and tasks.\footnote{Tailoring stylometric features to each dataset/task would likely produce even higher results, and we encourage testing specialized feature sets for more representative performance.}

\label{sec:attribute-analysis}

\paragraph{CISR and StyleDistance models perform best at content independence.} StyleDistance, CISR, and mStyleDistance perform the best at content-independence (i.e., the distractor variant of the order alignment tasks). This is expected as they were all trained similarly with hard negatives to improve content independence. Across models, though, there is still a lot of room for improvement on content-independence.

\paragraph{STAR is most sensitive to style across stylistics attributes.} STAR performs the best at recovering distinctive patterns across ``objects of study'' (e.g., genre, time, registers) beyond authorial styles. This might be influenced by its being one of few models that was trained on several different domains (i.e., social media, blogs, and books). %
Including more diverse training datasets (and potentially training tasks, cf.~\citealp{wegmann_survey_2026}) might thus be a promising direction for more generalizability.

\subsection{Per-application analysis} \label{sec:per-application-analysis}

To answer questions such as ``Which model is best for AI-text detection?'', we manually group datasets into four application-oriented clusters (\autoref{tab:per-application-clusters}): ATD, ATD where text has been modified to evade detection (Adv.), AV, and AR.
For AV we additionally report three sub-columns over the PAN style-change benchmarks ordered by how content-entangled the pairs are: Easy, Medium, and Hard, where Hard pairs do not share any semantic content.
The top five comprise three style-specific embeddings (STAR, LUAR-MUD, LUAR-CRUD) and two DeBERTa-v3 variants.
In ATD, DeBERTa-v3-large\footnote{While striking, this result is explained by the fact that DeBERTav3 adds a \emph{replaced token detection} objective, where a discriminator is trained to detect which tokens in a sequence were replaced by the MLM~\citep{he2023debertav}. This objective is none other than ATD in a pre-ChatGPT era (first DeBERTa version was published in 2021)!} leads in the standard setting, while in the adversarial setting, STAR and DeBERTa-v3-large perform best.
In AV and AR, the style-specific embedding models take the lead, suggesting that they are better suited for capturing the fine-grained nuances of everyday authors.
The Easy/Medium/Hard AV columns reveal a finer pattern. On the Easy and Medium sets (where topic can be exploited), both STAR and LUAR perform well.
However, in the Hard setting (where topic correlations cannot be exploited), CISR and StyleDistance perform best, suggesting that disentangling style from content matters for such applications.

\begin{table}[!t]
\centering
\small
\setlength{\tabcolsep}{4pt}
\begin{tabular}{cl ccc}
\toprule
& \textbf{Model} & \textbf{Pair Class.} & \textbf{Retrieval} & \textbf{Avg.} \\
\midrule
\multirow{5}{*}{\rotatebox[origin=c]{90}{\shortstack{\emph{Top 5}}}}
& deberta-v3-large & \textbf{74.03} & \textbf{47.77} & \textbf{60.90} \\
& StyleDistance & \underline{72.40} & \underline{44.45} & \underline{58.42} \\
& roberta-large & 71.91 & 44.02 & 57.96 \\
& CISR & 70.08 & 37.93 & 54.01 \\
& deberta-v3-base & 71.85 & 35.42 & 53.63 \\
\midrule
\multirow{8}{*}{\rotatebox[origin=c]{90}{\shortstack{\emph{Style}\\\emph{embeddings}}}}
& LUAR-MUD & 69.50 & 32.38 & 50.94 \\
& LUAR-CRUD & 72.37 & 30.07 & 51.22 \\
& mStyleDistance & 59.94 & 14.69 & 37.32 \\
& StyleDist. Synth. & 64.03 & 19.46 & 41.74 \\
& MSR & 69.43 & 28.38 & 48.90 \\
& STAR & 71.32 & 35.80 & 53.56 \\
& LISA & 62.67 & 26.70 & 44.68 \\
\bottomrule
\end{tabular}
\caption{Top 5 multilingual \texttt{STEB} results ($\times 100$).
Pair Class.\ averages AUC across 13 PAN13/14/15 AV datasets (Dutch, Greek, Spanish); Retrieval averages MRR across 4 PAN18 cross-domain AA datasets (French, Italian, Polish, Spanish).
\textbf{Bold} = best, \underline{underline} = 2nd best per column (across all models).
\label{tab:multilingual-results}}\vspace{-.5cm}
\end{table}

\subsection{Multilingual analysis} \label{sec:multilingual-analysis}

\texttt{STEB} includes non-English datasets across six languages (Dutch, French, Greek, Italian, Polish, Spanish) drawn from the PAN13/14/15 AV shared tasks and the PAN18 cross-domain AA shared task.
These datasets are excluded from the main \texttt{STEB} scores (\autoref{tab:overall-results} and ~\autoref{tab:attribute-clusters}) as most evaluated models are monolingual English encoders.
Here, we investigate whether multilingual style embeddings perform better than monolingual English style embeddings on these multilingual applications. 
We derive a single score by averaging across all dataset metrics and across the independent tasks (pair classification, retrieval).

\paragraph{Multilingual style-representations underperform their English counterparts}
\autoref{tab:multilingual-results} shows that the top-5 by multilingual average comprises two DeBERTa-v3 variants, RoBERTa-large, and two style embeddings (StyleDistance, CISR).
The explicitly multilingual style embeddings (MSR, mStyleDistance) \emph{do not} appear in the top five.
These results suggest that current multilingual training strategies are not yet effective, and as such, they remain a challenge for future work.\footnote{To stress-test this conclusion, we reproduce \citet{kim-etal-2025-leveraging}'s multilingual evaluation setup on the same PAN13/14/15 multilingual AV datasets and style embeddings. 
We find that \texttt{STEB}'s default chunk-and-pool strategy for handling large texts is the main driver. 
Under per-document fixed truncation, MSR achieves the top score, while with the chunk-and-pool strategy, StyleDistance and CISR perform best. 
These results underscore the necessity of shared evaluation protocols, since without them our conclusions would differ materially.
See~\autoref{sec:multilingual-truncation} for the full comparison.
}
The two style embeddings that do appear are notable for having been trained so as to separate semantic content from style, suggesting that this separation might be a useful strategy for the generalizability of style embeddings across languages.

\section{Conclusion}

We introduced \texttt{STEB}, a benchmark for evaluating embeddings across style-related NLP tasks, spanning \numTasks{} tasks, \numDatasets{} datasets, and \numLanguages{} languages.
Across \numModels{} models, we find that no single embedding is best.
Style embeddings that retain semantic signal lead on authorship-related applications where author and topic correlate, while content-disentangled style embeddings lead on content-controlled applications where topic shortcuts are removed.
On ATD, MLMs outperform style-specific embeddings, and on predicting second-order style features (e.g., genre, dialect, register, time), the authorship-focused STAR model leads.
Multilingual style embeddings underperform their English counterparts on the multilingual subset of \texttt{STEB}, highlighting cross-lingual style as an open problem.
We release \texttt{STEB} to enable consistent, reproducible comparisons and to lower the cost of measuring progress on style.
We welcome contributions from the community in improving the benchmark.

\section*{Limitations}

\paragraph{Leakage of datasets / contamination} Style embeddings are commonly trained and tested on similar datasets, often using datasets for training and testing with differing splits between publications. Representations might be tested on some data that they have seen during training. For example, the StyleDistance models were trained on one of the seven ``linguistic features'' datasets and on two of the eleven ``Content Independence'' datasets in \autoref{tab:attribute-clusters}. Excluding these shrinks the difference to CISR to $\approx$2 percentage points on content independence and make the synthetic version and CISR have almost the same overall performance. However, the overall ranking of the top models is preserved.

\paragraph{Bounded by dataset coverage} Both the empirical clustering in §\ref{sec:overall-results} and the attribute clustering in §\ref{sec:attribute-analysis} are bounded by what existing datasets expose.
In particular, most of the field's emphasis has been on authorship-related tasks, and as such other axes of analysis remain underexplored. 
We call on the community to create datasets that explore aspects other than authorship and to expand STEB with these additions.

\paragraph{Languages other than English remains underrepresented} It is a fact of the field that not as many non-English datasets are available.
As such, we inherit this limitation.
Moreover, only recently has there been an effort to extend style embedding models to multilingual settings~\citep{kim_leveraging_2025,qiu_mstyledistance_2025}. 

\bibliography{custom, zotero}
\appendix

\section*{Appendix}
\label{sec:appendix}

\section{Full aggregate results}\label{sec:full-cluster-results}

The full tables of results for the experiments in~\autoref{sec:overall-results},~\autoref{sec:per-application-analysis}, and~\autoref{sec:multilingual-analysis} are shown in~\autoref{tab:overall-results-full},~\autoref{tab:per-application-clusters-full}, and~\autoref{tab:multilingual-results-full}, respectively.

\begin{table*}[!t]
\centering
\small
\setlength{\tabcolsep}{4pt}
\begin{tabular}{l cccccc c}
\toprule
\textbf{Model} & \textbf{Clustering} & \textbf{All-to-All} & \textbf{Predefined} & \textbf{Order Align.} & \textbf{Retrieval} & \textbf{Probing} & \textbf{\texttt{STEB} score} \\
Num. Datasets & 53 & 24 & 26 & 11 & 4 & 4 & 122 \\ 
\midrule
\emph{Style specific} \\
LUAR-MUD & 20.03 & 62.33 & 72.33 & 20.81 & \underline{75.67} & 53.04 & 50.70 \\
LUAR-CRUD & 19.65 & 61.00 & \underline{73.12} & 19.80 & \textbf{77.64} & 53.75 & \textbf{50.82} \\
StyleDistance & 18.46 & 59.82 & 66.21 & \textbf{44.02} & 49.61 & 57.87 & 49.33 \\
StyleDistance (Synthetic) & 13.46 & 57.02 & 57.67 & \underline{42.76} & 31.53 & 56.69 & 43.19 \\
mStyleDistance & 7.35 & 53.83 & 53.71 & 31.38 & 20.04 & 45.96 & 35.38 \\
MSR & 16.67 & \underline{64.34} & 71.01 & 21.85 & 63.49 & 51.01 & 48.06 \\
CISR & 15.28 & 59.71 & 66.43 & 33.81 & 45.87 & 52.54 & 45.61 \\
STAR & \textbf{25.92} & \textbf{64.56} & \textbf{73.28} & 23.93 & 66.96 & 49.86 & \underline{50.75} \\
LISA & 12.74 & 60.91 & 62.40 & 16.48 & 42.96 & 46.38 & 40.31 \\
\midrule
\emph{Semantic embedders} \\
all-mpnet-base-v2 & 8.48 & 60.41 & 63.87 & 14.62 & 48.99 & 45.64 & 40.34 \\
GTE-base-en-v1.5 & 10.76 & 60.04 & 60.33 & 15.62 & 49.36 & 50.64 & 41.13 \\
GTE-large-en-v1.5 & 11.48 & 60.99 & 62.39 & 16.20 & 49.60 & 48.87 & 41.59 \\
E5-base-v2 & 9.28 & 60.40 & 63.42 & 15.26 & 49.62 & 52.13 & 41.69 \\
E5-large-v2 & 11.06 & 60.20 & 64.70 & 15.82 & 49.91 & 49.03 & 41.79 \\
BGE-base-en-v1.5 & 7.93 & 59.40 & 62.97 & 15.82 & 51.44 & 49.45 & 41.17 \\
BGE-large-en-v1.5 & 7.15 & 59.84 & 63.81 & 16.23 & 48.79 & 47.23 & 40.51 \\
Jina Embeddings v3 & 7.90 & 60.58 & 64.60 & 17.11 & 47.35 & 45.62 & 40.53 \\
Qwen3-Embedding-8B & 2.67 & 51.61 & 52.79 & 23.11 & 35.17 & 42.96 & 34.72 \\
\midrule
\emph{Masked language models} \\
BERT-large-cased & 12.02 & 60.99 & 67.11 & 21.57 & 58.88 & 55.16 & 45.96 \\
BERT-large-uncased & 11.18 & 60.44 & 66.92 & 21.18 & 60.23 & 52.48 & 45.41 \\
RoBERTa-base & 12.59 & 60.97 & 66.49 & 22.47 & 50.63 & \textbf{61.32} & 45.74 \\
RoBERTa-large & 13.21 & 60.97 & 68.79 & 20.71 & 61.07 & \underline{57.92} & 47.11 \\
DeBERTa-v3-base & 19.42 & 58.35 & 66.82 & 23.41 & 55.17 & 51.73 & 45.81 \\
DeBERTa-v3-large & \underline{21.44} & 59.37 & 70.45 & 24.23 & 63.24 & 53.41 & 48.69 \\
ModernBERT-base & 14.05 & 59.28 & 64.13 & 19.20 & 51.25 & 53.91 & 43.63 \\
ModernBERT-large & 16.58 & 60.19 & 65.38 & 20.56 & 52.17 & 54.06 & 44.82 \\
\midrule
\emph{Causal language models} \\
GPT-2 XL & 4.42 & 51.68 & 53.49 & 26.32 & 28.82 & 48.66 & 35.57 \\
OPT-1.3B & 3.83 & 52.40 & 55.59 & 24.08 & 39.07 & 45.82 & 36.80 \\
Qwen2-0.5B & 3.77 & 51.59 & 52.78 & 25.51 & 25.70 & 42.34 & 33.62 \\
Qwen3-0.6B-Base & 3.73 & 51.43 & 53.14 & 26.01 & 23.67 & 41.78 & 33.29 \\
Qwen3.5-0.8B-Base & 4.21 & 52.73 & 57.23 & 25.07 & 32.83 & 43.30 & 35.89 \\
Qwen3.5-2B-Base & 3.97 & 52.46 & 57.09 & 24.61 & 35.13 & 42.61 & 35.98 \\
Qwen3.5-4B-Base & 4.23 & 52.80 & 57.41 & 24.88 & 37.15 & 43.02 & 36.58 \\
\midrule
\emph{Pre-defined features} \\
Function words & 7.17 & 55.17 & 60.88 & 14.51 & 43.14 & 37.98 & 36.47 \\
TFIDF FineWeb 1-2-grams  & 6.78 & 57.18 & 64.45 & 13.72 & 50.95 & 55.74 & 41.47 \\
TFIDF FineWeb 1-3-grams & 6.37 & 56.97 & 64.45 & 13.67 & 49.00 & 55.51 & 40.99 \\
TFIDF Reddit 1-2-grams & 6.79 & 56.93 & 64.45 & 13.84 & 49.77 & 55.54 & 41.22 \\
TFIDF Reddit 1-3-grams & 6.52 & 56.68 & 64.25 & 13.84 & 50.26 & 55.34 & 41.15 \\
Stylometric & 9.67 & 56.88 & 60.48 & 25.87 & 39.52 & 53.30 & 40.95 \\
NeuroBiber & 8.24 & 55.10 & 57.27 & 14.13 & 9.06 & 51.64 & 32.57 \\
\bottomrule
\end{tabular}
\caption{Full \texttt{STEB} (operational) results ($\times 100$). \textbf{Bold} = best, \underline{underline} = 2nd best per column. Metrics: V-measure (Clustering), AUC (All-to-All and Predefined Pair Classification), distractor accuracy (Order Alignment), MRR (Retrieval), and average accuracy (Probing). Scores are macro-averaged across automatically discovered dataset clusters within each task, then averaged across tasks to produce the overall \texttt{STEB} score.
Num. Datasets counts the datasets contributing to each task; a dataset may contribute to more than one task.
}
\label{tab:overall-results-full}
\end{table*}

\begin{table*}[!t]
\centering
\footnotesize
\setlength{\tabcolsep}{4pt}
\begin{tabular}{l cc cccc cc}
\toprule
\textbf{Model} & \textbf{ATD} & \textbf{ATD (Adv.)} & \multicolumn{4}{c}{\textbf{Authorship Verification}} & \textbf{AR} & \textbf{Avg.} \\
\cmidrule(lr){4-7}
 &  &  & \textbf{Overall} & \textbf{Easy} & \textbf{Medium} & \textbf{Hard} &  &  \\
\midrule
\multicolumn{9}{l}{\textit{Top 5}} \\
\midrule
deberta-v3-large & \textbf{36.03} & \textbf{100.00} & 73.29 & 77.68 & 62.92 & 55.87 & 63.24 & \textbf{68.14} \\
STAR & 22.79 & \textbf{100.00} & \textbf{77.52} & 88.37 & \textbf{76.02} & 58.26 & 66.96 & \underline{66.82} \\
LUAR-MUD & 28.20 & 75.01 & \underline{76.54} & 86.84 & 73.62 & 61.10 & \underline{75.67} & 63.85 \\
deberta-v3-base & 29.55 & \underline{99.24} & 69.40 & 73.46 & 59.80 & 54.75 & 55.17 & 63.34 \\
LUAR-CRUD & 25.20 & 72.61 & 76.35 & 86.60 & \underline{75.92} & 61.23 & \textbf{77.64} & 62.95 \\
StyleDistance & \underline{30.10} & 70.08 & 70.44 & 75.52 & 69.17 & \underline{62.18} & 49.61 & 55.06 \\
ModernBERT-large & 26.39 & 60.69 & 68.10 & 85.02 & 64.09 & 50.75 & 52.17 & 51.84 \\
CISR & 22.78 & 64.33 & 71.57 & 79.39 & 70.74 & \textbf{65.56} & 45.87 & 51.14 \\
roberta-large & 23.13 & 35.31 & 71.34 & 85.49 & 66.43 & 52.61 & 61.07 & 47.71 \\
bert-large-uncased & 17.53 & 30.22 & 69.39 & 82.49 & 65.10 & 50.94 & 60.23 & 44.34 \\
bert-large-cased & 16.61 & 27.57 & 69.36 & 85.59 & 64.93 & 49.53 & 58.88 & 43.11 \\
ModernBERT-base & 24.85 & 27.94 & 66.71 & 83.86 & 61.03 & 49.89 & 51.25 & 42.69 \\
StyleDistance (Synthetic) & 22.76 & 44.03 & 59.00 & 64.82 & 55.28 & 52.62 & 31.53 & 39.33 \\
MSR & 17.38 & 0.12 & 73.88 & \textbf{88.92} & 73.43 & 55.99 & 63.49 & 38.72 \\
roberta-base & 24.83 & 8.29 & 69.02 & 84.16 & 64.37 & 50.31 & 50.63 & 38.19 \\
TFIDF Reddit 1-3-grams & 12.37 & 7.17 & 65.83 & 73.90 & 60.50 & 50.98 & 50.26 & 33.91 \\
TFIDF Reddit 1-2-grams & 11.93 & 4.64 & 66.20 & 75.72 & 60.34 & 50.79 & 49.77 & 33.13 \\
TFIDF FineWeb 1-2-grams & 12.77 & 0.81 & 66.44 & 77.03 & 60.69 & 51.03 & 50.95 & 32.74 \\
TFIDF FineWeb 1-3-grams & 13.36 & 1.25 & 66.28 & 75.46 & 61.09 & 51.12 & 49.00 & 32.47 \\
gte-base-en-v1.5 & 14.30 & 0.16 & 63.76 & 87.36 & 67.91 & 43.27 & 49.36 & 31.89 \\
Stylometric & 24.10 & 0.87 & 62.87 & 77.32 & 50.71 & 46.58 & 39.52 & 31.84 \\
e5-large-v2 & 9.48 & 0.20 & 67.16 & 88.54 & 68.08 & 48.21 & 49.91 & 31.69 \\
Function words & 14.25 & 5.20 & 63.10 & 68.40 & 58.25 & 52.30 & 43.14 & 31.42 \\
LISA & 11.61 & 6.06 & 65.03 & 84.41 & 66.71 & 47.86 & 42.96 & 31.41 \\
e5-base-v2 & 9.33 & 0.11 & 65.97 & 87.83 & 67.96 & 46.38 & 49.62 & 31.26 \\
gte-large-en-v1.5 & 8.36 & 0.20 & 64.93 & 86.67 & 67.65 & 44.15 & 49.60 & 30.77 \\
bge-base-en-v1.5 & 6.13 & 0.01 & 65.05 & 86.64 & 65.66 & 43.76 & 51.44 & 30.66 \\
all-mpnet-base-v2 & 6.25 & 0.04 & 66.00 & 87.36 & 66.33 & 44.30 & 48.99 & 30.32 \\
jina-embeddings-v3 & 6.56 & 0.01 & 67.15 & \underline{88.83} & 68.93 & 44.80 & 47.35 & 30.27 \\
bge-large-en-v1.5 & 5.97 & 0.01 & 65.90 & 87.12 & 66.79 & 44.15 & 48.79 & 30.17 \\
Qwen3.5-4B-Base & 9.01 & 1.86 & 57.55 & 62.97 & 51.11 & 47.61 & 37.15 & 26.39 \\
opt-1.3b & 8.13 & 0.03 & 54.78 & 55.99 & 51.32 & 50.35 & 39.07 & 25.50 \\
Qwen3.5-2B-Base & 8.27 & 1.27 & 57.02 & 61.82 & 50.61 & 47.77 & 35.13 & 25.42 \\
Qwen3.5-0.8B-Base & 9.25 & 1.80 & 57.33 & 62.35 & 51.71 & 47.86 & 32.83 & 25.30 \\
mStyleDistance & 15.85 & 5.48 & 54.41 & 60.49 & 49.63 & 50.53 & 20.04 & 23.95 \\
Qwen3-Embedding-8B & 6.95 & 0.11 & 52.75 & 62.78 & 52.41 & 47.54 & 35.17 & 23.74 \\
gpt2-xl & 12.52 & 0.09 & 52.84 & 53.91 & 51.43 & 49.49 & 28.82 & 23.57 \\
NeuroBiber & 20.34 & 3.18 & 59.02 & 71.14 & 55.94 & 52.30 & 9.06 & 22.90 \\
Qwen2-0.5B & 8.29 & 0.38 & 52.94 & 61.94 & 50.58 & 48.08 & 25.70 & 21.83 \\
Qwen3-0.6B-Base & 8.29 & 0.34 & 53.29 & 61.53 & 50.27 & 47.91 & 23.67 & 21.39 \\
\bottomrule
\end{tabular}
\caption{Scores across application clusters, sorted in descending order.
ATD = AI-Text Detection, AR = Authorship Retrieval. 
\textbf{Authorship Verification}: \emph{Overall} aggregates a superset of datasets (PAN13--15, PAN20--21, Enron, and the PAN22--26 style-change benchmarks), while \emph{Easy}/\emph{Medium}/\emph{Hard} are computed over the PAN22--26 style-change subset only---they do not average to the \emph{Overall} column. 
To avoid double-counting, only \emph{Overall} contributes to \emph{Avg.} 
\textbf{Bold} = best, \underline{underline} = 2nd best per column (across all models). Each column reports the macro-averaged score ($\times 100$) within the corresponding manual cluster. \label{tab:per-application-clusters-full}}
\end{table*}

\begin{table*}[!t]
\centering
\small
\setlength{\tabcolsep}{4pt}
\begin{tabular}{l ccc}
\toprule
\textbf{Model} & \textbf{Pair Class.} & \textbf{Retrieval} & \textbf{Avg.} \\
\midrule
deberta-v3-large & \textbf{74.03} & \textbf{47.77} & \textbf{60.90} \\
StyleDistance & \underline{72.40} & \underline{44.45} & \underline{58.42} \\
roberta-large & 71.91 & 44.02 & 57.96 \\
bert-large-cased & 71.33 & 39.11 & 55.22 \\
CISR & 70.08 & 37.93 & 54.01 \\
deberta-v3-base & 71.85 & 35.42 & 53.63 \\
STAR & 71.32 & 35.80 & 53.56 \\
roberta-base & 66.01 & 39.70 & 52.86 \\
bert-large-uncased & 69.36 & 34.92 & 52.14 \\
LUAR-CRUD & 72.37 & 30.07 & 51.22 \\
LUAR-MUD & 69.50 & 32.38 & 50.94 \\
TFIDF Reddit 1-2-grams & 63.02 & 38.52 & 50.77 \\
TFIDF Reddit 1-3-grams & 62.61 & 37.34 & 49.98 \\
MSR & 69.43 & 28.38 & 48.90 \\
TFIDF FineWeb 1-3-grams & 61.94 & 34.53 & 48.24 \\
TFIDF FineWeb 1-2-grams & 62.42 & 31.98 & 47.20 \\
e5-large-v2 & 67.05 & 24.60 & 45.83 \\
e5-base-v2 & 66.20 & 24.99 & 45.59 \\
ModernBERT-base & 69.09 & 20.55 & 44.82 \\
LISA & 62.67 & 26.70 & 44.68 \\
Stylometric & 67.38 & 19.37 & 43.37 \\
StyleDistance (Synthetic) & 64.03 & 19.46 & 41.74 \\
bge-base-en-v1.5 & 62.86 & 18.90 & 40.88 \\
ModernBERT-large & 66.95 & 13.80 & 40.37 \\
jina-embeddings-v3 & 64.02 & 16.23 & 40.12 \\
all-mpnet-base-v2 & 60.04 & 19.17 & 39.61 \\
bge-large-en-v1.5 & 62.93 & 14.61 & 38.77 \\
mStyleDistance & 59.94 & 14.69 & 37.32 \\
gte-large-en-v1.5 & 62.78 & 10.16 & 36.47 \\
Qwen3.5-0.8B-Base & 56.83 & 15.68 & 36.25 \\
Qwen3.5-2B-Base & 55.72 & 16.41 & 36.07 \\
Qwen3.5-4B-Base & 55.93 & 14.52 & 35.23 \\
Function words & 54.25 & 15.16 & 34.71 \\
gte-base-en-v1.5 & 58.53 & 10.41 & 34.47 \\
opt-1.3b & 55.30 & 11.55 & 33.43 \\
NeuroBiber & 54.98 & 10.59 & 32.79 \\
Qwen3-Embedding-8B & 54.56 & 10.51 & 32.54 \\
Qwen3-0.6B-Base & 54.64 & 10.41 & 32.53 \\
Qwen2-0.5B & 53.43 & 9.41 & 31.42 \\
gpt2-xl & 48.75 & 12.55 & 30.65 \\
\bottomrule
\end{tabular}
\caption{Full Multilingual \texttt{STEB} results ($\times 100$), sorted in descending order.
Pair Class.\ averages AUC across 13 PAN13/14/15 authorship verification datasets (Dutch, Greek, Spanish); Retrieval averages MRR across 4 PAN18 cross-domain AA datasets (French, Italian, Polish, Spanish).
\textbf{Bold} = best, \underline{underline} = 2nd best per column (across all models).
\label{tab:multilingual-results-full}}
\end{table*}

\begin{table*}[!t]
    \centering
    \small
    \setlength{\tabcolsep}{3pt}
    \begin{tabular}{l cccccc c c c c}
    \toprule
    & \multicolumn{7}{c}{\textbf{Object of Study}} & \textbf{Ling. Feat.} & \textbf{Content-Ind.} & \textbf{Avg.} \\
    \cmidrule(lr){2-8} \cmidrule(lr){9-9} \cmidrule(lr){10-10}
    \textbf{Model} & \textbf{Genre} & \textbf{Register} & \textbf{Time} & \textbf{Demo.} & \textbf{Dialect} & \textbf{Idiolect} & \textbf{Avg.} &  & \textbf{Order Al. $^{\dagger}$} & \\
    Num. Datasets & 4 & 7 & 3 & 3 & 3 & 29 & 49 & 7 & 11 & 67 \\ 
\midrule
\emph{Style specific} \\
StyleDistance & 62.84 & 40.01 & 46.21 & 33.67 & \textbf{56.91} & 60.03 & 49.94 & 74.02 & \underline{44.23} & \textbf{56.06} \\
StyleDistance (Synth.) & 55.45 & 38.81 & 43.27 & 33.13 & \underline{55.64} & 45.26 & 45.26 & 70.87 & \textbf{46.10} & \underline{54.08} \\
CISR & 62.43 & 36.09 & 41.52 & 33.03 & 54.76 & 58.72 & 47.76 & 65.85 & 32.31 & 48.64 \\
STAR & \textbf{68.68} & 39.79 & \textbf{47.25} & \textbf{40.99} & 49.39 & 72.24 & \textbf{53.06} & 67.66 & 14.08 & 44.93 \\
mStyleDistance & 49.41 & 32.83 & 35.04 & 32.25 & 42.83 & 37.23 & 38.27 & 56.17 & 38.32 & 44.25 \\
LUAR-MUD & \underline{66.67} & 38.72 & 42.14 & 35.53 & 44.30 & \underline{76.11} & \underline{50.58} & 70.50 & 11.29 & 44.12 \\
LUAR-CRUD & 65.80 & 38.14 & 41.30 & 35.32 & 41.52 & \textbf{76.99} & 49.85 & 70.37 & 10.57 & 43.59 \\
MSR & 66.39 & 39.08 & 41.98 & \underline{37.05} & 41.52 & 68.69 & 49.12 & 66.52 & 12.35 & 42.66 \\
LISA & 56.89 & 37.95 & 37.38 & 32.95 & 27.19 & 54.00 & 41.06 & 64.02 & 7.10 & 37.39 \\
\midrule
\emph{Semantic embedders} \\
Qwen3-Embedding-8B & 39.57 & 33.88 & 37.26 & 26.92 & 27.05 & 43.96 & 34.77 & 57.68 & 24.10 & 38.85 \\
E5-base-v2 & 55.91 & 37.26 & 39.72 & 30.50 & 22.85 & 57.80 & 40.67 & 69.27 & 5.11 & 38.35 \\
E5-large-v2 & 57.58 & 37.70 & 40.81 & 31.38 & 22.16 & 58.53 & 41.36 & 67.54 & 5.22 & 38.04 \\
GTE-base-en-v1.5 & 53.62 & 36.51 & 39.44 & 31.76 & 23.10 & 56.56 & 40.17 & 65.77 & 5.31 & 37.08 \\
BGE-base-en-v1.5 & 54.69 & 36.67 & 36.96 & 30.75 & 23.10 & 58.25 & 40.07 & 65.79 & 5.22 & 37.03 \\
BGE-large-en-v1.5 & 56.15 & 37.31 & 38.23 & 29.71 & 22.47 & 57.35 & 40.20 & 65.54 & 5.28 & 37.01 \\
Jina Embeddings v3 & 56.24 & 37.51 & 39.77 & 31.10 & 23.74 & 57.25 & 40.94 & 63.80 & 5.53 & 36.75 \\
GTE-large-en-v1.5 & 57.09 & 36.67 & 41.20 & 31.38 & 24.01 & 57.26 & 41.27 & 63.40 & 5.45 & 36.71 \\
all-mpnet-base-v2 & 56.07 & 37.55 & 39.03 & 31.47 & 23.31 & 57.50 & 40.82 & 62.18 & 5.06 & 36.02 \\
\midrule
\emph{Masked language models} \\
DeBERTa-v3-large & 63.26 & 38.25 & \underline{46.94} & 35.37 & 43.37 & 68.26 & 49.24 & 69.56 & 18.72 & 45.84 \\
RoBERTa-base & 63.79 & \textbf{42.52} & 45.94 & 32.84 & 40.16 & 59.83 & 47.51 & \textbf{77.95} & 10.27 & 45.24 \\
RoBERTa-large & 65.84 & 40.01 & 46.43 & 33.07 & 42.07 & 66.21 & 48.94 & \underline{75.48} & 9.86 & 44.76 \\
DeBERTa-v3-base & 58.34 & 36.63 & 44.97 & 33.72 & 46.08 & 62.29 & 47.01 & 68.28 & 18.49 & 44.59 \\
BERT-large-cased & 66.24 & \underline{41.99} & 45.11 & 34.79 & 41.23 & 64.12 & 48.91 & 73.58 & 9.22 & 43.90 \\
ModernBERT-large & 63.99 & 39.77 & 44.29 & 32.75 & 42.28 & 60.13 & 47.20 & 73.38 & 10.35 & 43.64 \\
ModernBERT-base & 62.23 & 38.76 & 43.14 & 32.67 & 37.71 & 58.98 & 45.58 & 72.90 & 9.64 & 42.71 \\
BERT-large-uncased & 62.24 & 39.90 & 45.27 & 32.80 & 44.03 & 64.81 & 48.17 & 68.55 & 10.89 & 42.54 \\
\midrule
\emph{Causal language models} \\
Qwen3.5-0.8B-Base & 46.66 & 35.59 & 40.51 & 28.29 & 32.34 & 45.08 & 38.08 & 60.61 & 26.03 & 41.57 \\
Qwen2-0.5B & 44.76 & 36.01 & 40.57 & 28.24 & 34.66 & 39.32 & 37.26 & 60.46 & 26.76 & 41.49 \\
GPT-2 XL & 41.36 & 32.22 & 36.15 & 27.29 & 34.00 & 40.83 & 35.31 & 58.54 & 29.14 & 41.00 \\
Qwen3.5-2B-Base & 45.71 & 34.81 & 39.04 & 28.33 & 33.39 & 46.08 & 37.89 & 59.49 & 25.49 & 40.96 \\
Qwen3.5-4B-Base & 44.42 & 34.62 & 38.93 & 28.43 & 34.70 & 47.35 & 38.07 & 58.63 & 25.72 & 40.81 \\
Qwen3-0.6B-Base & 42.81 & 34.26 & 38.22 & 28.26 & 35.77 & 38.48 & 36.30 & 57.99 & 28.12 & 40.80 \\
OPT-1.3B & 43.20 & 33.19 & 38.79 & 28.76 & 30.68 & 46.92 & 36.92 & 57.85 & 24.29 & 39.69 \\
\midrule
\emph{Pre-defined features} \\
Stylometric & 54.60 & 38.59 & 34.43 & 32.44 & 35.49 & 51.19 & 41.12 & 67.36 & 24.89 & 44.46 \\
Function words & 51.00 & 35.58 & 43.56 & 31.56 & 27.06 & 53.12 & 40.31 & 63.21 & 9.52 & 37.68 \\
TFIDF Reddit 1-2-grams & 49.91 & 33.96 & 36.19 & 31.67 & 27.71 & 57.98 & 39.57 & 66.90 & 5.66 & 37.38 \\
TFIDF FineWeb 1-2-grams & 49.19 & 33.79 & 36.20 & 31.70 & 26.31 & 58.70 & 39.32 & 67.13 & 5.55 & 37.33 \\
TFIDF Reddit 1-3-grams & 48.90 & 33.75 & 36.37 & 31.84 & 25.33 & 58.05 & 39.04 & 66.55 & 5.62 & 37.07 \\
TFIDF FineWeb 1-3-grams & 48.49 & 33.55 & 36.30 & 31.75 & 24.81 & 57.64 & 38.76 & 66.68 & 5.47 & 36.97 \\
NeuroBiber & 47.31 & 32.38 & 34.49 & 31.46 & 28.07 & 34.04 & 34.62 & 64.11 & 7.99 & 35.58 \\
\bottomrule
\end{tabular}
\caption{Full \texttt{STEB} (definitional) results. 
    Results are grouped into three clusters after \citet{wegmann_survey_2026} and averaged: Object of Study (Genre, Register, Time, Demographics, Dialect, Idiolect), Linguistic Features, and Content-Independence. 
    The Avg. column under Object of Study reports the mean of the six objects of study. Models are sorted by overall Avg. (descending).
    The datasets used for each column are given in App.~\autoref{tab:dataset-style-score}. 
    Order Alignment$^{\dagger}$ uses the \emph{distractor} variant of the order-alignment task for the respective datasts and no other task for that dataset. The other columns use all available tasks for the included dataset and the \emph{acc} variant of order alignment.
    \textbf{Bold} = best, \underline{underline} = 2nd best per column (across all models).
}
\label{tab:attribute-clusters-all}
\end{table*}

\section{Task examples}

We display an example of the order alignment task in Figure \ref{fig:order-alignment-example}.

\begin{figure}[t]
    \centering
    \small
    \textbf{Task:} Align the order of set B to that of set A w.r.t.\ style: \\[0.5em]
   	    \begin{tabular}{p{20pt} p{60pt} p{10pt} p{60pt}}
              &  \hspace*{20pt}  \textbf{1.} & & \hspace*{20pt} \textbf{2.}   \\
             \textbf{Set A} 
                & \cellcolor{green!25} r u a fan of them or something? &                & \textcolor{black}{Are you one of their fans? \tikzmark{A2_end} } \\ 
            & & \\ \midrule
            & \\
             \textbf{Set B} 
                & \tikzmark{A2_top} \textcolor{gray}{\st{Oh, and also that young physician got an unflattering haircut}}
                & & \cellcolor{green!25}Oh yea and that young dr got a bad haircut  \\ 
            \end{tabular}
            \begin{tikzpicture}[overlay, remember picture]
                \draw [->,very thick, red]($(A2_end.east) - (1.6,0.2)$) -- ($(A2_top.east) - (-2.35,-0.3)$) ;
            \end{tikzpicture}\\[1.5em]
    \begin{flushleft}\hspace{1em}\textbf{Solution:} B2, B1\end{flushleft}
    \caption{\textbf{Order Alignment Example.} Set A is written in an informal and a formal style, respectively. Set B is written in the reverse stylistic order. The task is to reorder B to match A's style sequence. We aim for sets to only include sentences showing the same content. The ``distractor'' variant is signified by the modifications in grey and red. Figure was taken and slightly modified from \citet{wegmann-etal-2022-author}\label{fig:order-alignment-example}.}
\end{figure}

\newpage
\onecolumn
\section{Dataset details}\label{sec:dataset_details}

Table~\ref{tab:dataset-attribute-overview} summarizes which stylistic attribute each \texttt{STEB} dataset is intended to probe, using the seven categories around which this section is organized.
A check mark indicates the category the dataset is grouped under in our analyses; some datasets could plausibly fit additional categories (e.g.\ the Corpus of Diverse Styles spans dialect, time, and genre) but we keep a single assignment for clarity.

\begin{center}
\footnotesize
\setlength{\tabcolsep}{3pt}
\renewcommand{\arraystretch}{0.95}
\begin{longtable}{l ccccccc}
\caption{\texttt{STEB} datasets by intended stylistic attribute. Feat.\ = linguistic feature, Dial.\ = dialect, Reg.\ = register, Gen.\ = genre, Demo.\ = demographics, Auth.\ = authorship. Each row receives a single check mark for the bucket the dataset is grouped under in this appendix; a few datasets span multiple attributes but are assigned to a single bucket for clarity.
 \label{tab:dataset-attribute-overview}}\\
\toprule
\textbf{Dataset} & \textbf{Feat.} & \textbf{Dial.} & \textbf{Reg.} & \textbf{Gen.} & \textbf{Time} & \textbf{Demo.} & \textbf{Auth.} \\
\midrule
\endfirsthead
\toprule
\textbf{Dataset} & \textbf{Feat.} & \textbf{Dial.} & \textbf{Reg.} & \textbf{Gen.} & \textbf{Time} & \textbf{Demo.} & \textbf{Auth.} \\
\midrule
\endhead
\bottomrule
\endlastfoot
STEL\_feature                     & \checkmark &            &            &            &            &            &            \\
SynthSTEL\_feature                & \checkmark &            &            &            &            &            &            \\
StylePTB                          & \checkmark &            &            &            &            &            &            \\
probing\_amazon                   & \checkmark &            &            &            &            &            &            \\
probing\_blog                     & \checkmark &            &            &            &            &            &            \\
probing\_reddit                   & \checkmark &            &            &            &            &            &            \\
probing\_stackexchange            & \checkmark &            &            &            &            &            &            \\
\midrule
twitter\_aave\_sae                &            & \checkmark &            &            &            &            &            \\
endive                            &            & \checkmark &            &            &            &            &            \\
eWAVE                             &            & \checkmark &            &            &            &            &            \\
\midrule
STEL\_register                    &            &            & \checkmark &            &            &            &            \\
SynthSTEL\_register               &            &            & \checkmark &            &            &            &            \\
OneStopEnglishCorpus              &            &            & \checkmark &            &            &            &            \\
graded\_formality                 &            &            & \checkmark &            &            &            &            \\
ASSET                             &            &            & \checkmark &            &            &            &            \\
wikipedia\_politeness             &            &            & \checkmark &            &            &            &            \\
stackexchange\_politeness         &            &            & \checkmark &            &            &            &            \\
enron\_spam                       &            &            & \checkmark &            &            &            &            \\
sms\_spam                         &            &            & \checkmark &            &            &            &            \\
telegram-spam-ham                 &            &            & \checkmark &            &            &            &            \\
hate\_speech                      &            &            & \checkmark &            &            &            &            \\
hate\_speech\_and\_offensive\_language &       &            & \checkmark &            &            &            &            \\
\midrule
corpus-of-diverse-styles          &            &            &            & \checkmark &            &            &            \\
core                              &            &            &            & \checkmark &            &            &            \\
x\_genre                          &            &            &            & \checkmark &            &            &            \\
masc\_text\_genre                 &            &            &            & \checkmark &            &            &            \\
\midrule
parallel\_shakespeare             &            &            &            &            & \checkmark &            &            \\
ceeces                            &            &            &            &            & \checkmark &            &            \\
bible\_versions                   &            &            &            &            & \checkmark &            &            \\
\midrule
fce\_l1                           &            &            &            &            &            & \checkmark &            \\
blog\_age                         &            &            &            &            &            & \checkmark &            \\
blog\_gender                      &            &            &            &            &            & \checkmark &            \\
pastel\_age                       &            &            &            &            &            & \checkmark &            \\
pastel\_education                 &            &            &            &            &            & \checkmark &            \\
pastel\_gender                    &            &            &            &            &            & \checkmark &            \\
pastel\_ethnic                    &            &            &            &            &            & \checkmark &            \\
pastel\_politics                  &            &            &            &            &            & \checkmark &            \\
pastel\_tod                       &            &            &            &            &            & \checkmark &            \\
\midrule
pan13\_authorship\_verification\_english\_test                          &            &            &            &            &            &            & \checkmark \\
pan14\_authorship\_verification\_corpus1\_english\_essays\_test         &            &            &            &            &            &            & \checkmark \\
pan14\_authorship\_verification\_corpus1\_english\_novels\_test         &            &            &            &            &            &            & \checkmark \\
pan14\_authorship\_verification\_corpus2\_english\_essays\_test         &            &            &            &            &            &            & \checkmark \\
pan14\_authorship\_verification\_corpus2\_english\_novels\_test         &            &            &            &            &            &            & \checkmark \\
pan15\_authorship\_verification\_english\_test                          &            &            &            &            &            &            & \checkmark \\
pan20\_authorship\_verification\_test                                   &            &            &            &            &            &            & \checkmark \\
pan21\_authorship\_verification\_test                                   &            &            &            &            &            &            & \checkmark \\
enron\_authorship\_corpus                                               &            &            &            &            &            &            & \checkmark \\
pan18\_style\_change                           &            &            &            &            &            &            & \checkmark \\
pan22\_style\_change\_basic                    &            &            &            &            &            &            & \checkmark \\
pan22\_style\_change\_advanced                 &            &            &            &            &            &            & \checkmark \\
pan22\_style\_change\_sentence                 &            &            &            &            &            &            & \checkmark \\
pan23\_style\_change\_easy                     &            &            &            &            &            &            & \checkmark \\
pan23\_style\_change\_medium                   &            &            &            &            &            &            & \checkmark \\
pan23\_style\_change\_hard                     &            &            &            &            &            &            & \checkmark \\
pan24\_style\_change\_easy                     &            &            &            &            &            &            & \checkmark \\
pan24\_style\_change\_medium                   &            &            &            &            &            &            & \checkmark \\
pan24\_style\_change\_hard                     &            &            &            &            &            &            & \checkmark \\
pan25\_style\_change\_easy                     &            &            &            &            &            &            & \checkmark \\
pan25\_style\_change\_medium                   &            &            &            &            &            &            & \checkmark \\
pan25\_style\_change\_hard                     &            &            &            &            &            &            & \checkmark \\
pan26\_style\_change\_easy                     &            &            &            &            &            &            & \checkmark \\
pan26\_style\_change\_medium                   &            &            &            &            &            &            & \checkmark \\
pan26\_style\_change\_hard                     &            &            &            &            &            &            & \checkmark \\
pan18\_cross\_domain\_authorship\_attribution\_english                  &            &            &            &            &            &            & \checkmark \\
amazon                                                                  &            &            &            &            &            &            & \checkmark \\
fanfiction                                                              &            &            &            &            &            &            & \checkmark \\
stackexchange\_retrieval                                                &            &            &            &            &            &            & \checkmark \\
gede\_essay\_detection                                                  &            &            &            &            &            &            & \checkmark \\
machine\_text\_detection\_DetectRL\_arxiv                               &            &            &            &            &            &            & \checkmark \\
machine\_text\_detection\_DetectRL\_direct\_prompt                      &            &            &            &            &            &            & \checkmark \\
machine\_text\_detection\_DetectRL\_writing\_prompt                     &            &            &            &            &            &            & \checkmark \\
machine\_text\_detection\_DetectRL\_xsum                                &            &            &            &            &            &            & \checkmark \\
machine\_text\_detection\_DetectRL\_yelp\_review                        &            &            &            &            &            &            & \checkmark \\
machine\_text\_detection\_DetectRL\_paraphrase\_attacks                 &            &            &            &            &            &            & \checkmark \\
machine\_text\_detection\_DetectRL\_perturbation\_attacks               &            &            &            &            &            &            & \checkmark \\
machine\_text\_detection\_DetectRL\_prompt\_attacks                     &            &            &            &            &            &            & \checkmark \\
machine\_text\_detection\_M4\_arxiv                                     &            &            &            &            &            &            & \checkmark \\
machine\_text\_detection\_M4\_peerread                                  &            &            &            &            &            &            & \checkmark \\
machine\_text\_detection\_M4\_reddit                                    &            &            &            &            &            &            & \checkmark \\
machine\_text\_detection\_M4\_wikihow                                   &            &            &            &            &            &            & \checkmark \\
machine\_text\_detection\_M4\_wikipedia                                 &            &            &            &            &            &            & \checkmark \\
machine\_text\_detection\_MAGE\_cmv                                     &            &            &            &            &            &            & \checkmark \\
machine\_text\_detection\_MAGE\_eli5                                    &            &            &            &            &            &            & \checkmark \\
machine\_text\_detection\_MAGE\_hswag                                   &            &            &            &            &            &            & \checkmark \\
machine\_text\_detection\_MAGE\_roct                                    &            &            &            &            &            &            & \checkmark \\
machine\_text\_detection\_MAGE\_sci\_gen                                &            &            &            &            &            &            & \checkmark \\
machine\_text\_detection\_MAGE\_squad                                   &            &            &            &            &            &            & \checkmark \\
machine\_text\_detection\_MAGE\_tldr                                    &            &            &            &            &            &            & \checkmark \\
machine\_text\_detection\_MAGE\_wp                                      &            &            &            &            &            &            & \checkmark \\
machine\_text\_detection\_MAGE\_xsum                                    &            &            &            &            &            &            & \checkmark \\
machine\_text\_detection\_MAGE\_yelp                                    &            &            &            &            &            &            & \checkmark \\
machine\_text\_detection\_PAN24\_news                                   &            &            &            &            &            &            & \checkmark \\
machine\_text\_detection\_PAN25\_26\_essays                             &            &            &            &            &            &            & \checkmark \\
machine\_text\_detection\_PAN25\_26\_fiction                            &            &            &            &            &            &            & \checkmark \\
machine\_text\_detection\_PAN25\_26\_news                               &            &            &            &            &            &            & \checkmark \\
machine\_text\_detection\_PAN25\_collaborative                          &            &            &            &            &            &            & \checkmark \\
\end{longtable}
\end{center}
\twocolumn

\subsection{Feature datasets}\label{app:data-feature}

\paragraph{STEL\_feature} added as \textit{order alignment} task. We split the publicly available STEL version \cite{wegmann-nguyen-2021-capture,wegmann-etal-2022-author} into registers (cf.~Section~\ref{app:data-register}) and linguistic features (i.e., number substitution, contraction or emoji use characteristic). STEL\_feature consist of $\approx400$ instances total. The STEL characteristics are publicly released. Note that in order alignment tasks we pair each style transfer pair with every other pair from the same style category---thus, we do not use the predefined pairings from STEL or SynthSTEL. 

\paragraph{SynthSTEL\_feature} added as \textit{order alignment} task. We split the SynthSTEL dataset \cite{patel_styledistance_2025} into 8 register (cf.~Section~\ref{app:data-register}) and 32 linguistic features (i.e., everything else, including features like All Lower Case / Proper Capitalization). We combine the train and test split, leading to 100 pairs per feature. SynthSTEL is released under the MIT license per its HuggingFace dataset card at \url{https://huggingface.co/datasets/StyleDistance/synthstel}.

\paragraph{StylePTB} added as \textit{order alignment} task. StylePTB~\citep{lyu_styleptb_2021} is a style-transfer benchmark with $\approx60k$ parallel pairs over 21 individual styles. 
We keep 15 and remove six out of the 21 dimensions as some change meaning (e.g., antonym replacement) or have no consistent style change within the same dimension (e.g., synonym replacement uses WordNet without clear stylistic commonalities in the selected synonyms like market replaced with marketplace and years with twelvemonths). 
Note that the distribution is heavily skewed: \texttt{to\_future} has $\approx7k$ pairs while \texttt{pp\_back\_to\_front} has only 7. 
To keep as many categories as possible, we combine the train/dev/test partitions. The text is provided in tokenized form: 
we do not detokenise. The StylePTB transformations are released under CC BY 4.0 (\url{https://github.com/lvyiwei1/StylePTB/blob/master/LICENSE}).

\paragraph{Probing}
added as \textit{probing} task across four domains (\texttt{probing\_amazon}, \texttt{probing\_blog}, \texttt{probing\_reddit}, \texttt{probing\_stackexchange}). 
We extract $46$ foundation-level linguistic features per document using the LFTK toolkit~\citep{leelee2023-lftk}: $12$ token-level counts (words, stop-words, punctuation, syllables, characters, plus token frequency norms) and $34$ POS-tag counts (total and unique counts for each of $17$ Universal Dependencies POS tags). 
Each feature is independently quantile-binned into discrete labels and the resulting per-feature classification task is solved with a small MLP probe on top of frozen embeddings.
The reported scores are macro-averaged across the $46$ features. 
The underlying source texts are sampled from publicly available Amazon product reviews~\citep{ni-etal-2019-justifying}, blog posts (via Kaggle \url{https://www.kaggle.com/datasets/rtatman/blog-authorship-corpus}), Reddit comments (via HuggingFace dataset \texttt{AnnaWegmann/StyleEmbeddingData}), and Stack Exchange posts, yielding $27,584$, $39,111$, $16,447$, and $9,253$ records respectively after binning and balancing.
The scripts used to create the probing datasets will be released publicly on our GitHub repository.

\subsection{Dialectal datasets}

\paragraph{Twitter AAE}  added as \textit{order alignment} task. We use a set of parallel constructed tweets written originally in AAE variety and rewritten in SAE by annotators\cite{groenwold_investigating_2020}. The dataset consists of $\approx 2k$ instances. It can be accessed via the EMNLP 2020 supplementary archive at \url{https://aclanthology.org/attachments/2020.emnlp-main.473.OptionalSupplementaryMaterial.zip}. There is no license file; we treat it as research-use only consistent with its release as paper supplementary material.

\paragraph{EnDive} added as \textit{clustering} and \textit{all-to-all pair classification} task. We add parallel sentences from the EnDive benchmark \cite{gupta_endive_2025} for Jamaican English, African American English,  Colloquial Singaporean English, Indian English and Chicano English across 12 NLU benchmarks\footnote{specifically logic\_bench\_yn, logic\_bench\_mcq svamp mbpp humaneval gsm8k, folio, boolq, copa, multirc, sst-2, wsc
} at \url{https://huggingface.co/datasets/abhaygupta1266/}. The dataset was created using few-shot prompting with examples from native speakers. License information is not provided, but we assume it to be permissive based on the fact that it is meant to be a public benchmarks freely shared on HuggingFace.

\paragraph{eWAVE} added as \textit{clustering} and \textit{all-to-all pair classification} task. We add dialect example sentences from the Electronic World Atlas of Varieties of English~\citep{kortmann2020ewave} Cross-Linguistic Data Formats (CLDF) release at \url{https://zenodo.org/records/17433568} and \url{https://github.com/cldf-datasets/ewave}. We include 53 varieties ranging from $\approx20$ to $\approx200$ instances. Varieties include Hong Kong English, Minx English, Bahamian Creole and Indian English. 
The CLDF release is licensed CC BY 4.0.

\subsection{Register datasets} \label{app:data-register}

\paragraph{STEL\_register} added as \textit{order alignment} tasks. We split the publicly available\footnote{The full formality dataset of 815 items is not publicly available, so we use the public version of STEL with 100 instances of two pair seach, see \url{https://github.com/nlpsoc/STEL/}} STEL version \cite{wegmann-nguyen-2021-capture,wegmann-etal-2022-author} into registers (i.e., formality and complexity dimension) and linguistic features (cf.~\ref{app:data-feature}). The registers consist of $\approx200$ instances each. 

\paragraph{SynthSTEL\_register} added as \textit{order alignment} task. We split the SynthSTEL dataset \cite{patel_styledistance_2025} %
into 8 register (e.g., formal tone, offensive language, sarcasm) and 32 linguistic features (cf.~Section \ref{app:data-feature}). Note that depending on ones definition of ``register'', some categories like positive sentiment expression might not be considered style \cite{wegmann_survey_2026}. We combine the train and test split, leading to 100 pairs per feature. SynthSTEL is released under the MIT license per its HuggingFace dataset card at \url{https://huggingface.co/datasets/StyleDistance/synthstel}.

\paragraph{OneStopEnglish} added as \textit{order alignment} task. We use the OneStopEnglish corpus \cite{vajjala_onestopenglish_2018} at {\url{https://github.com/nishkalavallabhi/OneStopEnglishCorpus/}}. 
It provides 189 parallel texts at three reading levels (beginner, intermediate, advanced) compiled from news articles rewritten by English teachers to suit the levels of the learners. %
It is shared with Creative Commons Attribution Share Alike 4.0 International, see \url{https://github.com/nishkalavallabhi/OneStopEnglishCorpus/blob/master/LICENSE.markdown}.

\begin{table}[t]
    \centering
    \begin{tabular}{@{}cl@{}}
        \toprule
        \textbf{Level} & \textbf{Example} \\
        \midrule
        1 & Lol your always so convincing. \\
        2 & Lol, you're always so convincing. \\
        3 & You're always so convincing. \\
        4 & You are always very convincing. \\
        5 & You are consistently persuasive. \\
        \bottomrule
    \end{tabular}
    \caption{Graded formality example.}
    \label{tab:graded-formality-examples}
\end{table}

\paragraph{Graded Formality Dataset} added as \textit{order alignment} task. From TweetEval \cite{barbieri_tweeteval_2020} we take the $\approx4k$ %
Tweets classified as non-offensive at \url{https://huggingface.co/datasets/cardiffnlp/tweet_eval/}. 
From SMS Spam \cite{almeida_contributions_2011}, we take the  $\approx4k$ SMS %
classified not spam at \url{ https://huggingface.co/datasets/ucirvine/sms_spam}. We use \texttt{gpt-5-mini-2025-08-07} rewrite the utterances more formally 4 times, see Figure \ref{fig:prompt}. Both datasets are publicly shared on HuggingFace. %
An example of the resulting dataset can found in Table~\ref{tab:graded-formality-examples}. 
An example of the resulting sentences are: ``Lol your always so convincing.'', ``Lol, you're always so convincing.'', ``You're always so convincing.'', ``You are always very convincing.'', ``You are consistently persuasive.''.

\paragraph{ASSET} added as \textit{order alignment} task. 
We use the ASSET corpus~\citep{alva-manchego_asset_2020}, a
multi-reference sentence-simplification benchmark covering  $\approx2k$ %
English Wikipedia source sentences with ten crowdsourced simplifications per sentence. To turn ASSET into an order-alignment task on the simple axis, we form one length-two ordered pair per source sentence by sampling one of its ten reference simplifications uniformly at random as the
simpler element and pairing it with the original Wikipedia sentence as the more-complex element. We combine the
\texttt{validation} and \texttt{test} splits on HuggingFace \url{https://huggingface.co/datasets/facebook/asset}. The HuggingFace card shows the license as CC BY SA 4.0.

\paragraph{wikipedia\_politeness}  added as \textit{clustering} and \textit{all-to-all pair classification} task. We add the Wikipedia Politeness Corpus \cite{danescu-niculescu-mizil_computational_2013} shared at \url{https://zissou.infosci.cornell.edu/convokit/datasets/wikipedia-politeness-corpus/}. %
The corpus consists of $\approx4k$ polite, neutral and impolite texts of Wikipedia editor requests. 
The corpus is shared with a CC BY license v4.0, see \url{https://convokit.cornell.edu/documentation/wiki_politeness.html}.

\paragraph{stackexchange\_politeness} added as \textit{clustering} and \textit{all-to-all pair classification} task. We add the  Stack Exchange Politeness Corpus of~\citet{danescu-niculescu-mizil_computational_2013} shared at \url{https://zissou.infosci.cornell.edu/convokit/datasets/stack-exchange-politeness-corpus/}. We use the same labels as wikipedia\_politeness. It consists of $\approx6k$  Stack Exchange requests. The corpus is shared with a CC BY license v4.0, see \url{https://convokit.cornell.edu/documentation/stack_politeness.html}

\subsection{Genre datasets}

\paragraph{Corpus of Diverse Styles} added as \textit{clustering},  \textit{all-to-all pair classification} and \textit{order alignment} task. We use the Corpus of Diverse Styles \citep{krishna_reformulating_2020} taken from HuggingFace at \url{https://huggingface.co/datasets/billray110/corpus-of-diverse-styles}. The original dataset totals $\approx400k$ records across 11 ``styles'' (i.e., {lyrics}, three decade slices of the Corpus of Historical American English, {Tweets}, African American English Tweets, the writing of James Joyce, the Bible, the Switchboard telephone-speech transcripts, English poetry, and Shakespeare). Note that while we could distribute the labels between dialect, time and genre, we assign everything to genre for simplicity. %
The code is MIT-licensed (\url{https://github.com/martiansideofthemoon/style-transfer-paraphrase/blob/master/LICENSE}). %

\paragraph{CORE} added as \textit{clustering} and \textit{all-to-all pair classification} task. We add the Corpus of Online Registers of English (CORE) \cite{laippala_register_2023} {as found at \url{https://github.com/TurkuNLP/CORE-corpus}}. We use the sub-labels as the relevant classes for pair classification, i.e., everything but the main labels \{IN, NA, HI, LY, SP, IP, ID, OP\}. If a document falls in several categories, we have it show up in several. We use documents from any file across the train, dev and test sets. This dataset was licensed under CC BY-SA 4.0.

\paragraph{x\_genre} added as \textit{clustering} and \textit{all-to-all pair classification} task.
We add the English slice of the X-GENRE genre-classification dataset~\citep{kuzman_automatic_2023} shared at \url{https://huggingface.co/datasets/TajaKuzmanPungersek/X-GENRE-text-genre-dataset}. X-GENRE merges FTD~\citep{sharoff_functional_2018} and CORE \cite{laippala_register_2023} into a joint 9-label schema using classification models. Compared to CORE, 15 sub-labels and their texts are discarded. The added dataset consists of $\approx2k$ English documents. We merge the train/dev/test partitions. X-GENRE is released under CC BY-SA 4.0 on HuggingFace.

\paragraph{MASC} added as \textit{clustering} and \textit{all-to-all pair classification} task. We add the Manually Annotated Sub-Corpus of American English (MASC) 3.0.0~\citep{ide_masc_2008,ide-etal-2013-masc-ldc} at \url{http://www.anc.org/data/masc/downloads}.  %
MASC is a corpus including texts from 19 genres in American English (e.g., debate transcripts, e-mail, jokes). %
We create 150 samples per genre. %
MASC is distributed without license or other restrictions, see \url{https://anc.org/data/masc/about/}.

\subsection{Time datasets}

\paragraph{Shakespeare} added as \textit{order alignment} task. We add the parallel Shakespeare corpus of \citet{xu_paraphrasing_2012} at \url{https://github.com/cocoxu/Shakespeare}. It includes labels for historical (i.e., Shakespearean) and contemporary English, providing $\approx$20k line-aligned pairs of originals and their  modern paraphrases for 17 plays. The license file only includes a request for citation at \url{https://github.com/cocoxu/Shakespeare/blob/master/README_LICENSE}, thus it is likely to be shared at least for research purposes.

\paragraph{CEECES} added as \textit{clustering} and \textit{all-to-all pair classification}. We add the Corpus of Early English Correspondence Extension Sampler (CEECES), parts 1 and 2~\citep{nevalainen2022ceeces1, nevalainen2022ceeces2}.  %
It consists of $\approx15k$ paragraphs from $\approx2.5k$ historical letters labeled by six 20-year periods from 1680-1800. 
Both Zenodo records are licensed under Creative Commons Attribution Non Commercial 4.0 International and Creative Commons Attribution Non Commercial No Derivatives 4.0 International at \url{https://zenodo.org/records/6411789} and \url{https://zenodo.org/records/5887101}.

\paragraph{Bible versions} added as \textit{clustering} and \textit{all-to-all pair classification} task. We add the parallel Bible corpus of~\citet{carlson_evaluating_2018} shared at \url{https://github.com/keithecarlson/StyleTransferBibleData}. The dataset comprises eight verse-aligned English Bible translations including, for example, the American Standard Version (from 1901) and the Kings James Version (from 1611). %
We remove verses that are identical (ignoring whitespacing and casing). The dataset includes no specific license, but all included Bibles are considered public domain and the associated paper was released with the Creative Commons Attribution License.

\subsection{Demographic datasets}

\paragraph{fce\_l1} added as \textit{clustering} and \textit{all-to-all pair classification} task for distinguishing different native language speakers. We add the the publicly released subset of the Cambridge Learner Corpus's First Certificate in English exam corpus (CLC FCE) \citep{yannakoudakis_new_2011} at \url{https://ilexir.co.uk/datasets/index.html}. %
It consists of $\approx2.5k$ responses for $\approx1.2k$ English Exams by students of 16 different native languages. The dataset is released with a non-commercial research and educational license.

\paragraph{blog\_age, blog\_gender} added as \textit{clustering} and \textit{all-to-all pair classification} tasks.  We add the Blog Authorship Corpus \cite{schler_effects_2006} via HuggingFace at \url{https://huggingface.co/datasets/barilan/blog_authorship_corpus}. We use the validation split to construct age (i.e., age buckets of 10s, 20s and 30s) and gender labeled texts. %
The corpus is meant for non-commercial research use, see \url{https://u.cs.biu.ac.il/~koppel/BlogCorpus.htm}.

\paragraph{PASTEL\_age, PASTEL\_education, PASTEL\_gender, PASTEL\_ethnic, PASTEL\_politics} added as \textit{clustering} and \textit{all-to-all pair classification} tasks. We add PASTEL~\citep{kang_male_2019} corpus at \url{https://github.com/dykang/PASTEL}. 
It consists of $\approx8k$ five-sentence stories written by crowd annotators. Texts are labeled with demographic information of annotators, including six age, eight education, three gender and, nine ethnic and three politics labels.
There is no explicit license information provided, but since the dataset is released as a benchmark, we assume at least a intended research-purpose license. 

\begin{table*}[th!]
    \centering\scriptsize
    \begin{tabular}{p{2cm}p{8cm}p{4cm}}
    \toprule
    \textbf{Task Type} & \textbf{Example} & \textbf{Solution}  \\
    \midrule
    \textbf{Order Alignment}
         & \textbf{Task:} Align the order of the set B to that of set A w.r.t.\ style: \newline
           \begin{tabular}[t]{@{}p{3.5cm}@{\hspace{0.5em}}|@{\hspace{0.5em}}p{3.5cm}@{}}
           \textbf{Set A} & \textbf{Set B} \\
           santa was to fat, and the woman was driving. & Santa was excessively overweight while the woman was driving. \\
           They cannot see anything in the beginning. & Well, they probably can't see anything at first. \\
           \end{tabular}
           & \mbox{}\newline\newline B2 \newline B1 \\ \midrule
    \textbf{Pair Classification}
         & \textbf{Task:} Are the two texts written in the same style? \newline
           \begin{tabular}[t]{@{}p{3.5cm}@{\hspace{0.5em}}|@{\hspace{0.5em}}p{3.5cm}@{}}
           \textbf{Text A} & \textbf{Text B} \\
           Santa was excessively overweight while the woman was driving. & santa was to fat, and the woman was driving.
           \end{tabular}
           & \mbox{}\newline\newline different style \\ \midrule
    \textbf{Clustering}
         & \textbf{Task:} Cluster the texts by style: \newline
           \begin{tabular}[t]{@{}p{7cm}@{}}
           \textbf{Text} \\
           T1. Round him peered Lenehan. \\
           T2. O lead me onward to the loneliest shade, \\
           T3. All Star Classic Game 1 Orlando 09 Game 1 - West Coast vs East \\
           T4. ( Goes up and comes down to center, shrieking and laughing. \\
           \end{tabular}
           & \mbox{}\newline\newline T1 $\to$ joyce \newline T2 $\to$ poetry \newline T3 $\to$ tweets \newline T4 $\to$ coha\_1890 \\ \midrule
    \textbf{Authorship Retrieval}
         & \textbf{Task:} Retrieve the target written by the same author as the query: \newline
           \begin{tabular}[t]{@{}p{7cm}@{}}
           \textbf{Query} \\
           Q. Bought as a Christmas gift but the case is nice \\[0.4em]
           \textbf{Targets} \\
           T1. Very well made and very loud! Feelnmich safer having it in my hunting bag! \\
           T2. Perfect for any animal lover - it says it all. The decal came shipped really nicely and shipped fast -
           \end{tabular}
           & \mbox{}\newline\newline T1 (author 100066) \\ \midrule
    \textbf{Probing}
         & \textbf{Task:} Predict each stylistic feature label from the text: \newline
           \begin{tabular}[t]{@{}p{7cm}@{}}
           \textbf{Text} \\
           Want to get one of these: But don't have enough money. Alas. \$249 for the smallest one seems a little steep.… \\[0.4em]
           \textbf{Features (which quintile does the feature fall into?)} \\
           \texttt{n\_adj}=1, \texttt{n\_verb}=0, \texttt{t\_word}=0, \texttt{t\_syll}=0, \dots\ \textit{(42 more features)}
           \end{tabular}
           & \mbox{}\newline\newline \texttt{n\_adj} $\to$ 1 \newline \texttt{n\_verb} $\to$ 0 \newline \texttt{t\_word} $\to$ 0 \newline \texttt{t\_syll} $\to$ 0 \\
    \bottomrule
    \end{tabular}
    \caption{\textbf{Examples of each \texttt{STEB} Tasks.
    } \label{tab:STEB-examples}}
\end{table*}

\subsection{Authorship datasets}

\paragraph{PAN Authorship Verification} added as \textit{predefined pair classification} task. 
We use the official test sets of the PAN authorship-verification shared tasks: PAN13~\citep{Juola2013OverviewOT}, PAN14~\citep{pan2014}, PAN15~\citep{stamatatos_overview_2015}, PAN20~\citep{pan2020}, and PAN21~\citep{pan2021}. 
PAN13, PAN14, and PAN15 contain multiple languages, including Greek, and Spanish in PAN13, Dutch, Greek, and Spanish in PAN14 and PAN15.
PAN20 and PAN21 are large-scale cross-domain fanfiction verification benchmarks, we subsample each to a balanced 500-pair subset (250 same-author, 250 different-author) with a fixed seed for reproducibility.
All corpora are available at Zenodo, and the download links can be found in the PAN website: \url{https://pan.webis.de/data.html}.
All PAN datasets are available for research use.

\paragraph{PAN Style Change} added as \textit{predefined pair classification} task. 
We use the publicly datasets of the PAN Style Change Detection shared tasks: PAN18~\citep{stamatatos_overview_2018}, PAN22~\citep{bevendorff_overview_2022}, PAN23~\citep{bevendorff_overview_2023}, PAN24~\citep{pan2024}, PAN2025~\citep{pan2025}, and PAN26~\citep{pan2026_preprint}.
For each of the aforementioned style cahnge datasets, consecutive paragraphs or sentence pairs (depending on the datasets granularity) are taken to be same- or different-author trials.
PAN23--26 ship three difficulty levels (\texttt{easy}, \texttt{medium}, \texttt{hard}) where each subsequent difficulty level has more topic overlap.
The PAN26 raw pairs are heavily imbalanced (up to $\approx$96\% same-author in the medium split).
We apply per-document stratified down-sampling that keeps all different-author pairs and matches the same-author count. 
All corpora are available at Zenodo, and the download links can be found in the PAN website: \url{https://pan.webis.de/data.html}.
All PAN datasets are available for research use under the CC BY 4.0 license.

\paragraph{Enron Authorship Corpus}~\citep{halvani2017enron} added as \textit{predefined pair classification} task. 
We add the 80-author Enron e-mail authorship-verification corpus at \url{https://prod-dcd-datasets-public-files-eu-west-1.s3.eu-west-1.amazonaws.com/f0527105-2774-423e-80ca-cc692b70b6cb}. 
Each verification case is comprised of $5$ documents, where $4$ documents come from one author and $1$ comes from another.
We concatenate the $4$ documents together separated by a double newline so as to create two pairs of text for the verification trail. 
The corpora is available for research use under the CC BY 4.0 licence.

\subsection{AI-text detection datasets}\label{app:data-mtd}

Standard ATD variants use per-generator labels (human plus one label per LLM).
Adversarial variants use binary human-vs-LLM labels because the question is whether embeddings still separate human from machine text after evasion techniques have been applied.

\paragraph{GEDE} added as a \textit{clustering} task for distinguishing human and the various LLMs. The Generative Essay Detection in Education (GEDE) dataset \cite{gehring2025} is a collection of three academic essay datasets (Argument Annotated Essays (AAE), PERSUADE 2.0, and British Academic Written English (BAWE)) plus machine-generated versions of those essays at various levels of contribution using GPT-4o-mini and Llama-3.3-70b-Instruct. There are a total of 916 human essays and 12,703 LLM essays. The dataset is provided under a CC BY-NC-SA 4.0 DEED Attribution-NonCommercial-ShareAlike 4.0 International license at \url{https://github.com/lukasgehring/Assessing-LLM-Text-Detection-in-Educational-Contexts}.

\paragraph{DetectRL}~\citep{wu2024detectrl} added as five standard machine-text-detection datasets (\texttt{DetectRL\_\{arxiv, direct\_prompt, writing\_prompt, xsum, yelp\_review\}} and three adversarial datasets (\texttt{DetectRL\_\{paraphrase, perturbation, prompt\}\_attacks}, binary human-vs-LLM). Source: \url{https://github.com/NLP2CT/DetectRL}.

\paragraph{M4}~\citep{wang-etal-2024-m4} added as five clustering datasets over English-only domains: \texttt{arxiv}, \texttt{peerread}, \texttt{reddit}, \texttt{wikihow}, and \texttt{wikipedia}.
Each domain contains data from $5$ different LLMs (Cohere, davinci, ChatGPT, Dolly-v2, and BloomZ), but we skip BloomZ due to the problematic nature of the data where parts of the prompt are in the ``generated" text.
Source: \url{https://github.com/mbzuai-nlp/M4}.
The data is available for research use and has previously been used in SemEval 2024 (\url{https://github.com/mbzuai-nlp/SemEval2024-task8/}).

\paragraph{MAGE}~\citet{li2024mage} added as ten clustering datasets over the HuggingFace dataset \texttt{yaful/MAGE}: \texttt{cmv}, \texttt{eli5}, \texttt{hswag}, \texttt{roct}, \texttt{sci\_gen}, \texttt{squad}, \texttt{tldr}, \texttt{wp}, \texttt{xsum}, and \texttt{yelp}. 
Each domain's records are labelled by generator model only, collapsing across generation methods (continuation, specified, topical), giving 27 unique machine labels plus \texttt{human} per domain. 
The data is available for public use in HuggingFace (\texttt{yaful/MAGE}). 

\paragraph{PAN24 Generative Authorship} added as \textit{clustering} and \textit{all-to-all pair classification} task. 
We use the test partition of the PAN24 Generative Authorship Detection~\citep{pan2024} task. 
All PAN datasets are available for research use under the CC BY 4.0 license.

\paragraph{PAN25/26 Generative AI Detection (Task 1)} added as three clustering datasets (\texttt{PAN25\_26\_\{essays, fiction, news\}}). 
The PAN25 and PAN26 editions of Task 1~\citep{bevendorff_overview_2025, pan2026_preprint} share the same underlying validation data.
We split it by the \texttt{genre} field. 
Each record is labelled by its \texttt{model} field, yielding a multi-class clustering task with human plus all participating LLMs per genre. 
All PAN datasets are available for research use under the CC BY 4.0 license.

\paragraph{PAN25 Human-AI Collaborative Text Classification} added as \textit{clustering} task. 
The PAN25 Task 2 development set~\citep{bevendorff_overview_2025} labels each text with one of six human-AI collaboration categories (e.g.\ \emph{fully human-written}, \emph{human-initiated, then machine-continued}, \emph{machine-initiated, then human-continued}). 
We perform the clustering across these different labels.
All corpora are available at Zenodo, and the download links can be found in the PAN website: \url{https://pan.webis.de/data.html}.
All PAN datasets are available for research use under the CC BY 4.0 license.

\subsection{Authorship retrieval datasets}\label{app:data-retrieval}

\paragraph{PAN18 Cross-Domain Authorship Attribution}~\citep{pan2018} added as \textit{retrieval} task. 
Each problem provides a set of candidate authors with one or more known texts and a set of unknown query texts to be attributed. 
We concatenate each candidate's known texts into a single target document and add each unknown text as a query with the corresponding ground-truth label. 
The PAN18 dataset also contains data in French, Italian, Polish, and Spanish.
All PAN datasets are available for research use under the CC BY 4.0 license.

\paragraph{Amazon and Reddit} added as \textit{retrieval} task. 
We use the same retrieval test splits as~\citet{rivera_soto_learning_2021}, subsampling each dataset to $1000$ query and target pairs to make the evaluation of all models tractable within our compute and timing constraints (the original datasets contain over $100,000$ queries and targets).
We re-distribute the subsampled test splits for research purposes only. 

\paragraph{StackExchange} added as \textit{retrieval} task.
StackExchange data is readily available for use from: \url{https://archive.org/download/stackexchange}.
It is licensed by CC BY-SA 3.0 which allows research use.
We collect a small sample of StackExchange data spanning $500$ authors who contribute to any community at random.
The retrieval setup is similar to~\citet{rivera_soto_learning_2021}, where multiple documents serve as a query and multiple documents serve as a target.

\section{TF-IDFngrams model details}\label{app:tfidf}
While many approaches compare the frequencies of particular n-grams, simply counting their presence in documents does not reveal broader contextual information, such as how common the n-gram is in general. %
Therefore, n-grams can be weighted using their Term Frequency-Inverse Document Frequency (TF-IDF), which considers how often an n-gram appears in a particular document compared to how rare it is in a corpus overall. This measurement increases the weight of rare words, which might be distinguishing of an author, and reduces the weight of common words. The authorship verification baseline for the PAN competition \cite{pan2023}, for instance, uses TF-IDF-weighted character 4-grams. 

We experimented with fitting the vectorizer to a few diverse datasets---Corpus of Diverse Styles \citep{krishna_reformulating_2020}, Reddit Million User Dataset \cite{khan_deep_2021}, and FineWeb \cite{penedo2024}---but found performance to be similar for all. After experimenting with various $n$ values for each of these models, we found that character 3-5-grams, token 1-2-grams, and POS tag 1-2-grams worked best overall. Based on these findings, for the main paper, we chose a TFIDFngrams model fit to a 10 billion token sample of FineWeb \cite{penedo2024}, a large collection of cleaned English web data from CommonCrawl, with these n-gram settings.~\autoref{tab:overall-results-full} through~\autoref{tab:attribute-clusters-all} show results for a few variations of TF-IDF n-gram models based on the dataset it was trained on (Reddit, FineWeb) and the chosen $n$ (1-2, 1-3) for the token n-grams and POS tag n-grams. The character n-gram values were not decreased because very short character n-grams occur extremely often but do not carry much discriminative information; this makes the feature matrix larger and sparser, increases computational cost, and can introduce noise that hurts generalization.

\begin{table*}[h!]
    \centering
    \small
    \setlength{\tabcolsep}{3pt}
    \renewcommand{\arraystretch}{0.95}
    \begin{tabular}{l cccccc c c}
    \toprule
    & \multicolumn{6}{c}{\textbf{Object of Study}} &  \textbf{Ling.\ Feat.} & \textbf{Content Independence}  \\
    \cmidrule(lr){2-7} \cmidrule(lr){8-8} \cmidrule(lr){9-9}
    \textbf{Dataset} & \textbf{Gen.} & \textbf{Register} & \textbf{Time} & \textbf{Demo.} & \textbf{Dialect} & \textbf{Idiolect} &  & \textbf{Order Alignment$^{\dagger}$} \\
    \midrule
    CORE                       & \checkmark &            &            &            &            &            &            &            \\
    x\_genre                   & \checkmark &            &            &            &            &            &            &            \\
    MASC                       & \checkmark &            &            &            &            &            &            &            \\
    Corpus of Diverse Styles   & \checkmark &            &            &            &            &            &            & \checkmark \\
    \midrule
STEL\_register             &            & \checkmark &            &            &            &            &            & \checkmark \\
SynthSTEL\_register        &            & \checkmark &            &            &            &            &            & \checkmark \\
Graded Formality           &            & \checkmark &            &            &            &            &            & \checkmark \\
OneStopEnglish             &            & \checkmark &            &            &            &            &            & \checkmark \\
ASSET                      &            & \checkmark &            &            &            &            &            & \checkmark \\
wikipedia\_politeness      &            & \checkmark &            &            &            &            &            &            \\
stackexchange\_politeness  &            & \checkmark &            &            &            &            &            &            \\
\midrule
Bible versions             &            &            & \checkmark &            &            &            &            &            \\
CEECES                     &            &            & \checkmark &            &            &            &            &            \\
Shakespeare                &            &            & \checkmark &            &            &            &            & \checkmark \\
\midrule
Blog                       &            &            &            & \checkmark &            &            &            &            \\
FCE L1                     &            &            &            & \checkmark &            &            &            &            \\
PASTEL                     &            &            &            & \checkmark &            &            &            &            \\
\midrule
EnDive                     &            &            &            &            & \checkmark &            &            &            \\
Twitter AAE/SAE            &            &            &            &            & \checkmark &            &            & \checkmark        \\
eWAVE                      &            &            &            &            & \checkmark &            &            &            \\
\midrule
Enron authorship corpus    &            &            &            &            &            & \checkmark &            &            \\
PAN13 AV                   &            &            &            &            &            & \checkmark &            &            \\
PAN14 AV                   &            &            &            &            &            & \checkmark &            &            \\
PAN15 AV                   &            &            &            &            &            & \checkmark &            &            \\
PAN20 AV                   &            &            &            &            &            & \checkmark &            &            \\
PAN21 AV                   &            &            &            &            &            & \checkmark &            &            \\
PAN18 SC                   &            &            &            &            &            & \checkmark &            &            \\
PAN22 SC                   &            &            &            &            &            & \checkmark &            &            \\
PAN23 SC                   &            &            &            &            &            & \checkmark &            &            \\
PAN24 SC                   &            &            &            &            &            & \checkmark &            &            \\
PAN25 SC                   &            &            &            &            &            & \checkmark &            &            \\
PAN26 SC                   &            &            &            &            &            & \checkmark &            &            \\
PAN18 cross-domain AA      &            &            &            &            &            & \checkmark &            &            \\
amazon                     &            &            &            &            &            & \checkmark &            &            \\
fanfiction                 &            &            &            &            &            & \checkmark &            &            \\
stackexchange\_retrieval   &            &            &            &            &            & \checkmark &            &            \\
\midrule
STEL\_feature              &            &            &            &            &            &            & \checkmark & \checkmark \\
SynthSTEL\_feature         &            &            &            &            &            &            & \checkmark & \checkmark \\
StylePTB                   &            &            &            &            &            &            & \checkmark & \checkmark \\
probing\_amazon            &            &            &            &            &            &            & \checkmark &            \\
probing\_blog              &            &            &            &            &            &            & \checkmark &            \\
probing\_reddit            &            &            &            &            &            &            & \checkmark &            \\
probing\_stackexchange     &            &            &            &            &            &            & \checkmark &            \\
\bottomrule
\end{tabular}
\caption{Datasets used for each cluster of the \texttt{STEB} Score (definitional). 
    This mirrors the three clusters of \autoref{tab:attribute-clusters} and \autoref{tab:attribute-clusters-all} after \citet{wegmann_survey_2026}: Object of Study (Genre, Register, Time, Demographics, Dialect, Idiolect), Linguistic Features (Ling.\ Feat.), and Content-Independence (Order Alignment task). 
    Idiolect includes authorship verification (PAN style-change and PAN authorship-verification editions, plus Enron) and authorship retrieval (PAN18 cross-domain AA English, Amazon, fanfiction, stackexchange\_retrieval). 
    PAN rows are collapsed across tasks and difficulty levels (AV = authorship verification, SC = style change, AA = authorship attribution); %
    Order Alignment$^{\dagger}$ uses the \emph{distractor} variant of the order-alignment task for the respective datasts and no other task for that dataset. The other columns use all available tasks for the included dataset and the \emph{acc} variant of order alignment.\label{tab:dataset-style-score}}
\end{table*}

\section{Embeddings vs. prompting}\label{app:small-retrieval}

\paragraph{In practice: Why not just use prompting?}To demonstrate this concretely, we compare prompting GPT-5.2 against LUAR-CRUD on authorship retrieval, using a relatively small corpus of 100 query and 100 target collections of 16 sentences each,  created with Amazon Reviews data (\autoref{app:small-retrieval}).
The dataset is a subsample of the \texttt{STEB} Amazon retrieval dataset described in~\autoref{app:data-retrieval}.
As no established prompts exist for retrieval (likely due to scaling issues, cf.~\autoref{app:small-retrieval}), we adapt the linguistically informed prompts created for authorship verification by \citet{huang_can_2024}, with minor grammatical corrections. We provide the full prompt in App.~\autoref{fig:prompt}.
Table \ref{tab:retrieval-llm} reports the same performance measures on retrieval as \texttt{STEB} (Section~\ref{sec:steb}). %
We additionally report tera floating point operations (TFLOPs) and API costs in US dollars (GPT-5.2 only). %
LUAR-CRUD, a model from 2021, outperforms GPT-5.2 released in 2025 while being at least 500$\times$ more efficent than causal models with 1B+ parameters. %

\begin{figure*}[ht]
\begin{tcolorbox}[title=System Message]
Respond with a JSON object with two elements:\\
\{\\
\quad "analysis": Reasoning behind your answer.\\
\quad "answer": A list of all candidate IDs (integers 0 to N-1) sorted from most to least
            likely to be written by the same author as the query text. The list must contain every
            candidate ID exactly once.\\
\}
\end{tcolorbox}
 
\begin{tcolorbox}[title=User Prompt]
You are given a set of texts written by one unknown person (the query author)
and several sets of candidate texts written by several known authors (the candidate authors).
Rank all candidate sets of texts by how likely each was written by the query author.
Analyze the writing style only, disregarding the differences in topic
and content. Base your reasoning on linguistic features such as phrasal verbs,
modal verbs, punctuation, rare words, affixes, quantities, humor, sarcasm,
typographical errors, and misspellings.
 
\medskip\noindent
Query texts:\\
\quad {[16 QUERY DOCUMENTS UNKNOWN AUTHOR]}
 
\medskip\noindent
Candidates (rank all $N$ by likelihood of matching the query author):\\
Candidate 0:\\
\quad [16 TARGET DOCUMENTS AUTHOR 0]\\
Candidate 1:\\
\quad [16 TARGET DOCUMENTS AUTHOR 1]\\
\quad $\vdots$\\
Candidate $N{-}1$:\\
\quad [16 TARGET DOCUMENTS AUTHOR $N-1$]
\end{tcolorbox}
\caption{Prompt template for AA via LLM-based stylistic analysis.}
\label{fig:prompt}
\end{figure*}

\label{app:retrieval-scaling}
We display the prompt used with GPT-5.2 in Figure~\ref{fig:prompt}.
The lower efficiency of LLM models is unsurprising: authorship retrieval requires comparing a query against every target collection in the pool, which scales poorly with prompting approaches. As the target pool grows, it may even exceed the model's context window or lead to large costs in US dollars or FLOPs.\footnote{This parallels problems faced in information retrieval and RAG \cite{gao_retrieval-augmented_2024}.} To illustrate, a common Reddit authorship retrieval dataset \cite{khan_deep_2021} contains 120k authors with 16 texts each, %
requiring approximately 5.1M tokens per retrieval query, %
which is considerably larger than GPT-5.2's context window of 400k.\footnote{See \url{https://developers.openai.com/api/docs/models/gpt-5.2}}
Even if the context window was not restricted, retrieval on the 120k dataset would require about 20 billion TFLOPs for a 1 billion parameter causal model and 40k TFLOPS for models like LUAR.

\section{Multilingual truncation analysis}\label{sec:multilingual-truncation}

The main-text multilingual ranking (\autoref{tab:multilingual-results}) is computed under \texttt{STEB}'s default protocol, which segments long inputs into sentence-boundary-aware chunks and mean-pools the chunk embeddings (\autoref{sec:chunking}).
To isolate the effect of this evaluation choice, we reproduce the multilingual evaluation setup of \citet{kim-etal-2025-leveraging}: the same PAN13/14/15 authorship verification datasets (Dutch, Greek, Spanish), the same pre-defined pair classification protocol, the same style embeddings, and the same AUC metric.
The only change is that each document is truncated to the model's maximum length rather than chunked and pooled as in \texttt{STEB}.
\autoref{tab:multilingual-truncation} compares the two protocols.
Under truncation, MSR moves from rank $4$ to rank $1$, while without truncation it ranks $4$th, underperforming the style embeddings trained on English data.
These results underscore the necessity of shared evaluation protocols, as without them our conclusions would differ materially.

\begin{table}[H]
\centering
\small
\setlength{\tabcolsep}{4pt}
\begin{tabular}{l cc cc}
\toprule
& \multicolumn{2}{c}{\textbf{Chunk \& Pool}} & \multicolumn{2}{c}{\textbf{Truncate}} \\
\cmidrule(lr){2-3} \cmidrule(lr){4-5}
\textbf{Model} & \textbf{AUC} & \textbf{Rank} & \textbf{AUC} & \textbf{Rank} \\
\midrule
StyleDistance     & \textbf{72.40} & 1 & 58.40 & 4 \\
CISR   & 70.08          & 2 & 60.31 & 2 \\
LUAR-MUD          & 69.50          & 3 & 58.52 & 3 \\
MSR               & 69.43          & 4 & \textbf{67.51} & 1 \\
mStyleDistance    & 59.94          & 5 & 56.88 & 5 \\
\bottomrule
\end{tabular}
\caption{Mean AUC ($\times 100$) on the PAN13/14/15 multilingual AV datasets used by \citet{kim-etal-2025-leveraging}, under \texttt{STEB}'s default chunk-and-pool protocol versus per-document truncation. \textbf{Bold} = best per protocol.
\label{tab:multilingual-truncation}}
\end{table}

\section{Chunking large documents}\label{sec:chunking}

\begin{figure*}[h!]
\begin{tikzpicture}[
    node distance=2cm, %
    >=Stealth,
    doc/.style={draw, rectangle, minimum width=2.5cm, minimum height=4cm, fill=gray!10, align=center, text width=2cm, font=\bfseries},
    chunk/.style={draw, rectangle, minimum width=2.2cm, minimum height=0.8cm, fill=blue!10, align=center, font=\small},
    embed/.style={draw, rectangle, rounded corners, minimum width=0.6cm, minimum height=1.5cm, fill=orange!10},
    op/.style={draw, circle, inner sep=4pt, fill=white, font=\Large\bfseries},
    label_text/.style={font=\bfseries\large}
]

    \node[doc] (longdoc) {Long Input Document\\[0.5em] \normalfont\scriptsize (e.g. 2048 tokens)};

    \node[chunk, right=3cm of longdoc.north, anchor=north, yshift=-0.3cm] (chunk1) {Chunk 1 \\ \scriptsize [0-512)};
    \node[chunk, below=0.8cm of chunk1] (chunk2) {Chunk 2 \\ \scriptsize [512-1024)};
    \node[chunk, below=0.8cm of chunk2] (chunk3) {Chunk $N$ \\ \scriptsize (Remainder)};

    \node[label_text, above=0.5cm of chunk1] {Segmentation};

    \coordinate (doc_right) at (longdoc.east);
    \draw[->, thick, gray] (doc_right |- chunk1.west) -- (chunk1.west);
    \draw[->, thick, gray] (doc_right |- chunk2.west) -- (chunk2.west);
    \draw[->, thick, gray] (doc_right |- chunk3.west) -- (chunk3.west);

    \node[embed, right=2cm of chunk1] (emb1) {};
    \node[embed, right=2cm of chunk2] (emb2) {};
    \node[embed, right=2cm of chunk3] (emb3) {};

    \node[label_text, above=0.5cm of emb1] {Encodings};

    \draw[->, thick] (chunk1) -- node[above, font=\scriptsize] {Encoder} (emb1);
    \draw[->, thick] (chunk2) -- node[above, font=\scriptsize] {Encoder} (emb2);
    \draw[->, thick] (chunk3) -- node[above, font=\scriptsize] {Encoder} (emb3);
    
    \node[below=0.2cm of emb3, font=\scriptsize] {Embeddings};

    \node[op, right=2.5cm of emb2] (pool) {$\frac{1}{N}\sum$};
    
    \node[label_text, above=0.5cm of pool|-emb1.north] {Mean Pooling};

    \draw[->, dashed, thick] (emb1.east) -- (pool);
    \draw[->, dashed, thick] (emb2.east) -- (pool);
    \draw[->, dashed, thick] (emb3.east) -- (pool);

    \node[embed, fill=green!10, right=2.5cm of pool] (final) {};
    \draw[->, thick] (pool) -- (final);

    \node[below=0.3cm of final, font=\bfseries] {Final Vector};
\end{tikzpicture}
\caption{Example of \texttt{STEB}s chunk-and-pool strategy.
In the example, it is assumed that the model's maximum context-length is $512$ tokens.
The long input document is chunked up into segments of $512$ tokens that respect sentence boundaries.
Each chunk is then embedded individually be the encoder, and finally we mean-pool across the chunks derive our final embedding.
}
\label{fig:chunking}
\end{figure*}
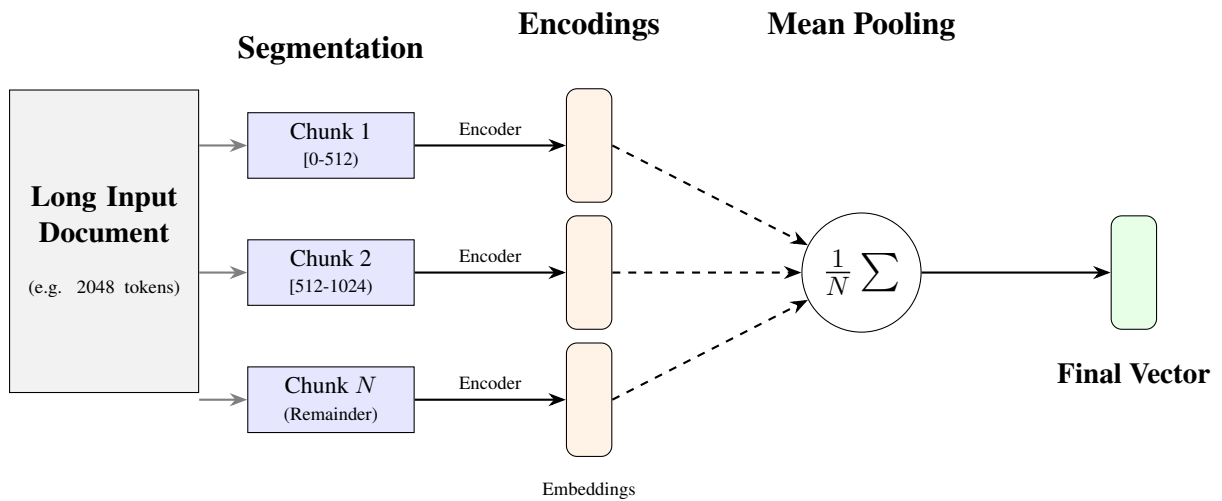

We demonstrate \texttt{STEB}s strategy for embedding large inputs into a single representation in~\autoref{fig:chunking}.
Note that this approach is fair to all encoders, as each one is able to observe the same amount of tokens regardless of what their token limits are.

\section{Use of AI Assistants}\label{app:use-of-ai}
We used AI-assistants (Claude Code, Cursor) to help us develop the STEB code-base.
That said, all the code underwent human-review, and we rigorously tested the code-base for correctness. 
We also used LLMs to give us ideas for rephrasing and shortening excerpts of the paper.

\section{Compute Requirements}\label{app:compute-requirements}
We ran each model on a single H100 or A100 40Gb GPU.
No single run took longer than 24hrs, and the largest model evaluated has 8B parameters.
We upper-bound our GPU expenditure at $960$ GPU hours ($24$ hours times $40$ models).
Please note that this is a very large over-estimate, most models took less than $8$ hours to run.

\onecolumn
\newpage
\section{Sample Counts per Dataset}\label{app:sample-counts}

We specify the number of classes, number of samples per class, and total number of texts embedding in~\autoref{tab:dataset-sample-counts}.

\begin{longtable}{@{}l r r r@{}}
\caption{Per-dataset sample counts in STEB. ``\#~Classes'' refers to 
  classes for clustering and pair-classification tasks, trials for 
  pre-defined pair classification, style groups for order alignment, 
  unique author/item ids for retrieval, and probing features for probing. 
  ``Samples/Class'' is the per-class evaluation budget. 
  ``Total'' is the total number of texts the embedder is asked to encode.}
\label{tab:dataset-sample-counts} \\
\toprule
Dataset & \# Classes & Samples/Class & Total \\
\midrule
\endfirsthead
\multicolumn{4}{c}{\tablename\ \thetable{} -- continued} \\
\toprule
Dataset & \# Classes & Samples/Class & Total \\
\midrule
\endhead
\midrule \multicolumn{4}{r}{Continued on next page} \\
\endfoot
\bottomrule
\endlastfoot
\multicolumn{4}{l}{\textbf{Clustering} (53 datasets)} \\
\midrule
bible\_versions & 8 & 200 & 1,600 \\
blog\_age & 3 & 200 & 600 \\
blog\_gender & 2 & 200 & 400 \\
ceeces & 6 & 200 & 1,200 \\
core & 38 & 25 & 950 \\
corpus-of-diverse-styles & 11 & 200 & 2,200 \\
eWAVE & 53 & 29 & 1,537 \\
endive & 6 & 200 & 1,200 \\
enron\_spam & 2 & 200 & 400 \\
fce\_l1 & 15 & 25 & 375 \\
gede\_essay\_detection & 4 & 200 & 800 \\
hate\_speech & 2 & 200 & 400 \\
hate\_speech\_and\_offensive\_language & 3 & 200 & 600 \\
machine\_text\_detection\_DetectRL\_arxiv & 5 & 200 & 1,000 \\
machine\_text\_detection\_DetectRL\_direct\_prompt & 5 & 200 & 1,000 \\
machine\_text\_detection\_DetectRL\_paraphrase\_attacks & 2 & 200 & 400 \\
machine\_text\_detection\_DetectRL\_perturbation\_attacks & 2 & 200 & 400 \\
machine\_text\_detection\_DetectRL\_prompt\_attacks & 2 & 200 & 400 \\
machine\_text\_detection\_DetectRL\_writing\_prompt & 5 & 200 & 1,000 \\
machine\_text\_detection\_DetectRL\_xsum & 5 & 200 & 1,000 \\
machine\_text\_detection\_DetectRL\_yelp\_review & 5 & 200 & 1,000 \\
machine\_text\_detection\_M4\_arxiv & 6 & 200 & 1,200 \\
machine\_text\_detection\_M4\_peerread & 6 & 200 & 1,200 \\
machine\_text\_detection\_M4\_reddit & 6 & 200 & 1,200 \\
machine\_text\_detection\_M4\_wikihow & 5 & 200 & 1,000 \\
machine\_text\_detection\_M4\_wikipedia & 5 & 200 & 1,000 \\
machine\_text\_detection\_MAGE\_cmv & 28 & 60 & 1,680 \\
machine\_text\_detection\_MAGE\_eli5 & 28 & 73 & 2,044 \\
machine\_text\_detection\_MAGE\_hswag & 28 & 66 & 1,848 \\
machine\_text\_detection\_MAGE\_roct & 28 & 76 & 2,128 \\
machine\_text\_detection\_MAGE\_sci\_gen & 28 & 36 & 1,008 \\
machine\_text\_detection\_MAGE\_squad & 28 & 55 & 1,540 \\
machine\_text\_detection\_MAGE\_tldr & 28 & 57 & 1,596 \\
machine\_text\_detection\_MAGE\_wp & 28 & 70 & 1,960 \\
machine\_text\_detection\_MAGE\_xsum & 28 & 86 & 2,408 \\
machine\_text\_detection\_MAGE\_yelp & 28 & 42 & 1,176 \\
machine\_text\_detection\_PAN24\_news & 14 & 200 & 2,800 \\
machine\_text\_detection\_PAN25\_26\_essays & 13 & 25 & 325 \\
machine\_text\_detection\_PAN25\_26\_fiction & 12 & 25 & 300 \\
machine\_text\_detection\_PAN25\_26\_news & 14 & 59 & 826 \\
machine\_text\_detection\_PAN25\_collaborative & 6 & 200 & 1,200 \\
masc\_text\_genre & 19 & 150 & 2,850 \\
pastel\_age & 6 & 25 & 150 \\
pastel\_education & 8 & 25 & 200 \\
pastel\_ethnic & 9 & 25 & 225 \\
pastel\_gender & 3 & 55 & 165 \\
pastel\_politics & 3 & 200 & 600 \\
pastel\_tod & 5 & 200 & 1,000 \\
sms\_spam & 2 & 200 & 400 \\
stackexchange\_politeness & 3 & 200 & 600 \\
telegram-spam-ham & 2 & 200 & 400 \\
wikipedia\_politeness & 3 & 200 & 600 \\
x\_genre & 9 & 46 & 414 \\
\midrule
\multicolumn{4}{l}{\textbf{All-to-All Pair Classification} (24 datasets)} \\
\midrule
bible\_versions & 8 & 200 & 1,600 \\
blog\_age & 3 & 200 & 600 \\
blog\_gender & 2 & 200 & 400 \\
ceeces & 6 & 200 & 1,200 \\
core & 38 & 25 & 950 \\
corpus-of-diverse-styles & 11 & 200 & 2,200 \\
eWAVE & 53 & 29 & 1,537 \\
endive & 6 & 200 & 1,200 \\
enron\_spam & 2 & 200 & 400 \\
fce\_l1 & 15 & 25 & 375 \\
hate\_speech & 2 & 200 & 400 \\
hate\_speech\_and\_offensive\_language & 3 & 200 & 600 \\
masc\_text\_genre & 19 & 150 & 2,850 \\
pastel\_age & 6 & 25 & 150 \\
pastel\_education & 8 & 25 & 200 \\
pastel\_ethnic & 9 & 25 & 225 \\
pastel\_gender & 3 & 55 & 165 \\
pastel\_politics & 3 & 200 & 600 \\
pastel\_tod & 5 & 200 & 1,000 \\
sms\_spam & 2 & 200 & 400 \\
stackexchange\_politeness & 3 & 200 & 600 \\
telegram-spam-ham & 2 & 200 & 400 \\
wikipedia\_politeness & 3 & 200 & 600 \\
x\_genre & 9 & 46 & 414 \\
\midrule
\multicolumn{4}{l}{\textbf{Pre-Defined Pair Classification} (25 datasets)} \\
\midrule
enron\_authorship\_corpus & 80 & 2 & 160 \\
pan13\_authorship\_verification\_english\_test & 30 & 2 & 60 \\
pan14\_authorship\_verification\_corpus1\_english\_essays\_test & 100 & 2 & 200 \\
pan14\_authorship\_verification\_corpus1\_english\_novels\_test & 100 & 2 & 200 \\
pan14\_authorship\_verification\_corpus2\_english\_essays\_test & 200 & 2 & 400 \\
pan14\_authorship\_verification\_corpus2\_english\_novels\_test & 200 & 2 & 400 \\
pan15\_authorship\_verification\_english\_test & 500 & 2 & 1,000 \\
pan18\_style\_change & 1,492 & 2 & 2,984 \\
pan20\_authorship\_verification\_test & 500 & 2 & 1,000 \\
pan21\_authorship\_verification\_test & 500 & 2 & 1,000 \\
pan22\_style\_change\_advanced & 9,537 & 2 & 19,074 \\
pan22\_style\_change\_basic & 2,141 & 2 & 4,282 \\
pan22\_style\_change\_sentence & 22,105 & 2 & 44,210 \\
pan23\_style\_change\_easy & 2,826 & 2 & 5,652 \\
pan23\_style\_change\_hard & 4,112 & 2 & 8,224 \\
pan23\_style\_change\_medium & 7,013 & 2 & 14,026 \\
pan24\_style\_change\_easy & 2,471 & 2 & 4,942 \\
pan24\_style\_change\_hard & 4,131 & 2 & 8,262 \\
pan24\_style\_change\_medium & 4,592 & 2 & 9,184 \\
pan25\_style\_change\_easy & 10,247 & 2 & 20,494 \\
pan25\_style\_change\_hard & 10,648 & 2 & 21,296 \\
pan25\_style\_change\_medium & 12,759 & 2 & 25,518 \\
pan26\_style\_change\_easy & 71,108 & 2 & 142,216 \\
pan26\_style\_change\_hard & 54,141 & 2 & 108,282 \\
pan26\_style\_change\_medium & 24,406 & 2 & 48,812 \\
\midrule
\multicolumn{4}{l}{\textbf{Order Alignment} (11 datasets)} \\
\midrule
ASSET & 1 & 400 & 400 \\
OneStopEnglishCorpus & 1 & 600 & 600 \\
STEL\_feature & 3 & 200 & 600 \\
STEL\_register & 2 & 392 & 784 \\
StylePTB & 14 & 50 & 700 \\
SynthSTEL\_feature & 32 & 200 & 6,400 \\
SynthSTEL\_register & 8 & 200 & 1,600 \\
corpus-of-diverse-styles & 11 & 400 & 4,400 \\
graded\_formality & 2 & 1,000 & 2,000 \\
parallel\_shakespeare & 1 & 400 & 400 \\
twitter\_aave\_sae & 1 & 400 & 400 \\
\midrule
\multicolumn{4}{l}{\textbf{Retrieval} (4 datasets)} \\
\midrule
amazon & 500 & 195 & 97,787 \\
fanfiction & 500 & 86 & 43,411 \\
pan18\_cross\_domain\_authorship\_attribution\_english & 50 & 5 & 259 \\
stackexchange\_retrieval & 500 & 125 & 62,980 \\
\midrule
\multicolumn{4}{l}{\textbf{Probing} (4 datasets)} \\
\midrule
probing\_amazon & 46 & 27,584 & 27,584 \\
probing\_blog & 46 & 39,111 & 39,111 \\
probing\_reddit & 46 & 16,447 & 16,447 \\
probing\_stackexchange & 46 & 9,253 & 9,253 \\
\end{longtable}

\twocolumn

\end{document}